\documentclass{article}

\usepackage{arxiv}

\usepackage[utf8]{inputenc} 
\usepackage[T1]{fontenc}    
\usepackage{hyperref}       
\usepackage{url}            
\usepackage{amsfonts}       
\usepackage{nicefrac}       
\usepackage{microtype}      

\usepackage{graphicx}
\usepackage{natbib}
\usepackage{doi}

\usepackage{setspace}

\usepackage{amsmath}
\usepackage{amssymb,mathabx,bm}
\usepackage{textgreek}

\usepackage{algorithm,algorithmic}
\usepackage{graphicx}
\usepackage{caption}
\usepackage{subcaption}
\usepackage{float}
\usepackage{multirow}
\usepackage{soul}
\usepackage{accents}
\usepackage[normalem]{ulem}

\usepackage{adjustbox}
\usepackage{listings}
\usepackage[theme=grayscale]{jlcode}

\newfloat{lstfloat}{htbp}{lop}
\floatname{lstfloat}{Listing}
\floatstyle{plain}

\usepackage{tikz}
\usepackage{pgfplots}
\usetikzlibrary{intersections}
\usepgfplotslibrary{fillbetween}
\pgfplotsset{compat=1.5}
\usetikzlibrary{pgfplots.groupplots}
\usetikzlibrary{shapes.geometric,backgrounds}

\newcommand{\obs}[1]{\accentset{\circ}{#1}}

\DeclareMathOperator*{\argmin}{arg\,min}

\title{Reactive Message Passing for Scalable Bayesian Inference}

\date{\today}

\author{\name Dmitry Bagaev \email d.v.bagaev@tue.nl \\
  \addr Department of Electrical Engineering\\
  Eindhoven University of Technology \\
  Eindhoven, the Netherlands
  \AND
  \name Bert de Vries \email bert.de.vries@tue.nl \\
  \addr Department of Electrical Engineering\\
  Eindhoven University of Technology \& GN Hearing \\
  Eindhoven, the Netherlands
}

\author{Dmitry Bagaev\\
	Department of Electrical Engineering\\
	Eindhoven University of Technology\\
	Eindhoven, the Netherlands \\
	\texttt{d.v.bagaev@tue.nl} \\
	\And
	Bert de Vries \\
	Department of Electrical Engineering\\
	Eindhoven University of Technology \& GN Hearing\\
	Eindhoven, the Netherlands \\
	\texttt{bert.de.vries@tue.nl} \\
}

\renewcommand{\shorttitle}{Reactive Message Passing for Scalable Bayesian Inference}

\hypersetup{
pdftitle={Reactive Message Passing for Scalable Bayesian Inference},
pdfauthor={Dmitry Bagaev, Bert de Vries},
pdfkeywords={Bayesian Inference, Factor Graphs, Message Passing, Reactive Programming, Variational Inference},
}

\begin{document}
\maketitle

\begin{abstract}
	We introduce Reactive Message Passing (RMP) as a framework for executing  schedule-free, robust and scalable message passing-based inference in a factor graph representation of a probabilistic model. RMP is based on the reactive programming style that only describes how nodes in a factor graph react to changes in connected nodes. The absence of a fixed message passing schedule improves robustness, scalability and execution time of the inference procedure. We also present ReactiveMP.jl, which is a Julia package for realizing RMP through minimization of a constrained Bethe free energy. By user-defined specification of local form and factorization constraints on the variational posterior distribution, ReactiveMP.jl executes hybrid message passing algorithms including belief propagation, variational message passing, expectation propagation, and expectation maximisation update rules. Experimental results demonstrate the improved performance of ReactiveMP-based RMP in comparison to other Julia packages for Bayesian inference across a range of probabilistic models. In particular, we show that the RMP framework is able to run Bayesian inference for large-scale probabilistic state space models with hundreds of thousands of random variables on a standard laptop computer.
\end{abstract}


\section{Introduction}\label{section:introduction}

In this paper, we develop a \emph{reactive} approach to Bayesian inference on factor graphs. We provide the methods, implementation aspects, and simulation results of message passing-based inference realized by a reactive programming paradigm. Bayesian inference methods facilitate the realization of a very wide range of useful applications, but in our case, we are motivated by our interest in execution real-time Bayesian inference in state space models with data streams that potentially may deliver an infinite number of observations over an indefinite period of time.

The main idea of this paper is to combine message passing-based Bayesian inference on factor graphs with a reactive programming approach to build a foundation for an efficient, scalable, adaptable, and robust Bayesian inference implementation. Efficiency implies less execution time and less memory consumption than alternative approaches; scalability refers to running inference in large probabilistic models with possibly hundreds of thousands of random variables; adaptability implies real-time in-place probabilistic model adjustment, and robustness relates to protection against failing sensors and missing data. We believe that the proposed approach, which we call \emph{Reactive Message Passing} (RMP), will lubricate the transfer of research ideas about Bayesian inference-based agents to real-world applications. To that end, we have developed and also present ReactiveMP.jl, which is an open source Julia package that implements RMP on a factor graph representation of a probabilistic model. Our goal is that ReactiveMP.jl grows to support practical, industrial applications of inference in sophisticated Bayesian agents and hopefully will also drive more research in this area.

In Section~\ref{section:motivation}, we motivate the need for RMP by analysing some weaknesses of alternative approaches. Section \ref{section:background} reviews background knowledge on message passing-based inference on factor graphs and variational Bayesian inference as a constrained Bethe free energy optimisation problem. The rest of the paper discusses our main contributions:
\begin{itemize}
  \item In Section~\ref{section:rmp}, we present the idea of Reactive Message Passing as a way to perform event-driven reactive Bayesian inference by message passing in factor graphs. RMP is a surprisingly simple idea of combining two well-studied approaches from different fields: message passing-based Bayesian inference and reactive programming;
  \item In Section \ref{section:implementation}, we present an efficient and scalable implementation of RMP for automated Bayesian inference in the form of the ReactiveMP.jl package for the Julia programming language. We also introduce a specification language for the probabilistic model and inference constraints;
  \item In Section \ref{section:experiments}, we benchmark the ReactiveMP.jl on various standard probabilistic signal processing models and compare it with existing message passing-based and sampling-based Bayesian inference implementations. We show that our new implementation scales easily for conjugate state space models that include hundreds of thousands random variables on a regular MacBook Pro laptop computer. For these models, in terms of execution time, memory consumption, and scalability, the proposed RMP implementation Bayesian inference outperforms the existing solutions by hundreds orders of magnitude and takes roughly a couple of milliseconds or a couple of minutes depending on the size of a dataset and a number of random variables in a model. 
\end{itemize}

Finally, in Section \ref{section:discussion} we discuss work-in-progress and potential future research directions.

\section{Motivation}\label{section:motivation}

Open access to efficient software implementations of strong mathematical or algorithmic ideas often leads to sharply increasing advances in various practical fields. For example, backpropagation in artificial neural networks stems from at least the 1980's, but practical applications have skyrocketed in recent years due to new solutions in hardware and corresponding software implementations such as TensorFlow \citep{martin_abadi_tensorflow_2015} or PyTorch \citep{paszke_pytorch_2019}.

However, the application of Bayesian inference for real-world signal processing problems still remains a big challenge. If we consider an autonomous robot that tries to find its way in a new terrain, we would want it to reason about its environment in real-time as well as to be robust to potential failures in its sensors. Furthermore, the robot preferably has the ability to not only to adapt to new observations, but also to adjust its internal representation of the current environment in real-time. Additionally, the robot will have limited computational capabilities and should be energy-efficient. These issues form a very challenging barrier in the deployment of real-time Bayesian inference-based synthetic agents to real-world problems.

In this paper, our goal is to build a foundation for a new approach to efficient, scalable, and robust Bayesian inference in state space models and release an open source toolbox to support the development process. By efficiency, we imply the capability of performing real-time Bayesian inference with a limited computational and energy budget. By scalability, we mean inference execution to acceptable accuracy with limited resources, even if the model size and number of latent variables gets very large. Robustness is also an important feature, by which we mean that if the inference system is deployed in a real-world setting, then it needs to stay continually operational even if part of the system collapses. 

We propose a combination of message passing-based Bayesian inference on Forney-style Factor Graphs and the reactive programming approach, which, to our knowledge, is less well-known and new in the message passing literature. Our approach is inspired by the neuroscience community and the \textit{Free Energy Principle} \citep{friston_free-energy_2009} since the brain is a good example of a working system that already realizes real-time and robust Bayesian inference at a large scale for a small energy consumption budget. 

\paragraph{Message passing} 

Generative models for complex real-world signals such as speech or video streams are often described by highly factorised probabilistic models with sparse structure and few dependencies between latent variables. Bayesian inference in such models can be performed efficiently by message passing on the edges of factor graphs. A factor graph visualizes a factorised representation of a probabilistic model where edges represent latent variables and nodes represent functional dependencies between these variables. Generally, as the models scale up to include more latent variables, the fraction of direct dependencies between latent variables decreases and as a result the factor graph becomes sparser. For highly factorized models, efficient inference can be realized by message passing as it naturally takes advantage of the conditional independencies between variables.

Existing software packages for message passing-based Bayesian inference, such as ForneyLab.jl \citep{van_de_laar_forneylab.jl:_2018} and Infer.Net \citep{minka_infernet_2018}, have been implemented in an imperative programming style. Before the message passing algorithm in these software packages can be executed, a message passing \textit{schedule} (i.e., the sequential order of messages) needs to be specified \citep{elidan_residual_2012, bishop_pattern_2006}. In our opinion, an imperative coding style and reliance on a pre-specified message schedule might be problematic from a number of viewpoints. For example, a fixed pre-computed schedule requires a full analysis of the factor graph that corresponds to the model, and if the model structure adapts, for instance, by deleting a node, then we are forced to stop the system and create a new schedule. This restriction makes it hard to deploy these systems in the field, since either we abandon model structure optimization or we give up on the guarantee that the system is continually operational. In addition, in real-world signal processing applications, the data often arrives asynchronously and may have significantly different update rates in different sensory channels. A typical microphone has a sample rate of 44.1 kHz, whereas a typical video camera sensor has a sample rate of 30-60 Hz. These factors create complications since an explicitly sequential schedule-based approach requires an engineer to create different schedules for different data sources and synchronise them explicitly, which may be cumbersome and error-prone.

\paragraph{Reactive programming}

In this paper, we provide a fresh look on message passing-based inference from an implementation point of view. We explore the feasibility of using the reactive programming (RP) paradigm as a solution for the problems stated above. Essentially, RP supports to run computations by dynamically reacting to changes in data sources, thus eliminating the need for a pre-computed synchronous message update scheme. Benefits of using RP for different problems have been studied in various fields starting from physics simulations \citep{boussinot_reactive_2015} to neural networks \citep{busch_pushnet_2020} and probabilistic programming as well \citep{baudart_reactive_2019}. We propose a new reactive version of message passing framework, which we simply call \textit{Reactive Message Passing} (RMP). The new message passing-based inference framework is designed to run without any pre-specified schedule, autonomously reacts to changes in data, scales to large probabilistic models with hundreds of thousands of unknowns, and, in principle, allows for more advanced features, such as run-time probabilistic model adjustments, parallel inference execution, and built-in support for asynchronous data streams with different update rates.

To support further development we present our own implementation of the RMP framework in the form of a software package for the Julia programming language called ReactiveMP.jl. We show examples and benchmarks of the new implementation for different probabilistic models including a Gaussian linear dynamical system, a hidden Markov model and a non-conjugate hierarchical Gaussian filter model. More examples including mixture models, autoregressive models, Gaussian flow models, real-time processing, update rules based on the expectation propagation algorithm, and others are available in the ReactiveMP.jl repository on GitHub.

\section{Background}\label{section:background}
This section first briefly introduces message passing-based exact Bayesian inference on Forney-style Factor Graphs. Then, we extend the scope to approximate Bayesian inference by Variational Message Passing based on minimisation of the Constrained Bethe Free Energy.

\paragraph{Forney-style Factor Graphs} In our work, we use Terminated Forney-style Factor Graphs (TFFG) \citep{forney_codes_2001} to represent and visualise conditional dependencies in probabilistic models, but the concept of RMP should be compatible with all other graph-based model representations. For good introductions to Forney-style factor graphs, we refer the reader to \cite{loeliger_introduction_2004, korl_factor_2005}. 

\paragraph{Inference by Message Passing} 

We consider a probabilistic model with probability density function $p(s, y) = p(y|s)(s)$ where $y$ represents observations and $s$ represents latent variables. The general goal of Bayesian inference is to estimate the \emph{posterior} probability distribution $p(s|y=\hat{y})$. Often, it is also useful to estimate \emph{marginal} posterior distributions
\begin{equation}
    p(s_i|y=\hat{y}) = \iint p(s|y=\hat{y}) \mathrm{d}s_{\setminus i},
    \label{eq:marginal_posterior_distibutions}
\end{equation}
for the individual components $s_i$ of $s$. In the case of discrete states or parameters, the integration is replaced by a summation.

In high-dimensional spaces, the computation of \eqref{eq:marginal_posterior_distibutions} quickly becomes intractable, due to an exponential explosion of the size of the latent space \citep{bishop_pattern_2006}. The TFFG framework provides a convenient solution for this problem. As a simple example, consider the factorised distribution
\begin{equation}
    p(x, \theta|y=\hat{y}) \,\propto\, f(\theta)g(x, \theta)h(y, x)\cdot\delta(y - \hat{y}).
    \label{eq:simple_generative_model}
\end{equation}

\begin{figure}[h]
    \centering
    \includegraphics[width=0.95\columnwidth]{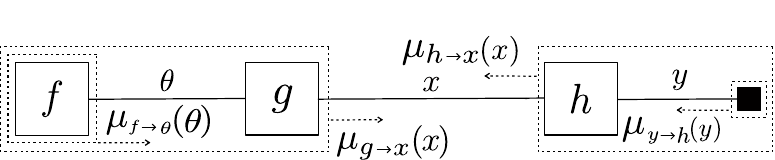}
    \caption{The message passing scheme for \eqref{eq:q_x_factorised_integrals}. Each nested integral can be interpreted as a message that flows between factor nodes on the TFFG. The messages are computed by the corresponding integrals in \eqref{eq:q_x_factorised_integrals}. }
    \label{fig:ffg_messages}
\end{figure}

This model can be represented by a TFFG shown in Fig.~\ref{fig:ffg_messages}. The underlying factorisation of $p$ allows the application of the algebraic distributive law, which simplifies the computation of \eqref{eq:marginal_posterior_distibutions} to a nested set of lower-dimensional integrals as shown in \eqref{eq:q_x_factorised_integrals}, which effectively reduces the exponential complexity of the computation to linear. In the corresponding TFFG, these nested integrals can be interpreted as messages that flow on the edges between nodes.

\begin{equation}\label{eq:q_x_factorised_integrals}
    \begin{gathered}
        p(x|y=\hat{y}) \, \propto \iint p(x, y, \theta)\cdot\delta(y - \hat{y}) \,\mathrm{d}y\,\mathrm{d}\theta = \overbrace{\int \underbrace{f(\theta)}_{\mu_{f \rightarrow \theta}(\theta)}g(x, \theta) \,\mathrm{d}\theta}^{\mu_{g \rightarrow x}(x)} \cdot \underbrace{\int h(y, x)\overbrace{\delta(y - \hat{y})}^{\mu_{y \rightarrow h}(y)}\mathrm{d}y }_{\mu_{h\rightarrow x}(x)}
    \end{gathered}
\end{equation}

The resulting marginal $p(x|y=\hat{y})$ in \eqref{eq:q_x_factorised_integrals} is simply equal to the product of two incoming (or colliding) messages on the same edge, divided by the normalisation constant
\begin{equation}
    p(x|y=\hat{y}) = \frac{\mu_{g \rightarrow x}(x) \cdot \mu_{h \rightarrow x}(x)}{\int \mu_{g \rightarrow x}(x) \cdot \mu_{h \rightarrow x}(x) \,\mathrm{d}x}.
    \label{eq:marginal_frac_norm}
\end{equation}

This procedure for computing a posterior distribution is known as the \textit{Belief Propagation} (BP) or \textit{Sum-Product} (SP) message passing algorithm and it requires only the evaluation of low-dimensional integrals over local variables. In some situations, for example when all incoming messages $\mu_{x_i \rightarrow f}(x_i)$ are Gaussian and the node function $f$ is linear, it is possible to use a closed-form analytical solutions for the messages. These closed-form formulas are also known as \textit{message update rules}. 

\paragraph{Variational Bayesian inference} Inference problems in practical models often involve computing messages in \eqref{eq:marginal_frac_norm} which are difficult to evaluate analytically. In these cases, we may resort to approximate Bayesian inference solutions with the help of variational Bayesian inference methods \citep{winn_variational_2003, dauwels_variational_2007}. In general, the variational inference procedure introduces a \emph{variational} posterior $q(s) \in \mathcal{Q}$ that acts as an approximating distribution to the Bayesian posterior $p(s|y=\hat{y})$. The set $\mathcal{Q}$ is called a \textit{variational family of distributions}. Typically, in variational Bayesian inference methods the aim is minimize the \textit{Variational Free Energy} (VFE) functional  

\begin{align}
F[q] &\triangleq \underbrace{\int q(s) \log \frac{q(s)}{p(s|y=\hat{y})}\mathrm{d}s}_{\mathrm{KL}\left[q(s)\;||\;p(s|y=\hat{y})\right]} \: - \log p(y=\hat{y})\,,\label{eq:VFE}\\
q^*(s) &=  \argmin_{q(s)\in \mathcal{Q}}F[q].\label{eq:variational_free_energy_optimisation}
\end{align}

In most cases, additional constrains on the set $\mathcal{Q}$ make the optimisation task \eqref{eq:variational_free_energy_optimisation} computationally more efficient than the belief propagation but only gives an approximate solution for \eqref{eq:marginal_posterior_distibutions}. The literature distinguishes two major types of constraints on $\mathcal{Q}$: form constraints and factorisation constraints \citep{senoz_variational_2021}. Form constraints force the approximate posterior $q(s)$ or its marginals $q_i(s_i)$ to be of a specific parametric functional form, e.g., a Gaussian distribution $q(s) = \mathcal{N}(s\,|\mu,\Sigma)$. Factorisation constraints on the posterior $q(s)$ introduce extra conditional independency assumptions that are not present in the generative model $p(y, s)$. 

\paragraph{Constrained Bethe Free Energy minimization} 

The VFE optimisation procedure in \eqref{eq:variational_free_energy_optimisation} can be framed as a message passing algorithm, leading to the so-called \textit{Variational message passing} (VMP) algorithm. A very large set of message passing algorithms, including BP and VMP, can also be interpreted as VFE minimization augmented with the \emph{Bethe factorization assumption} 
\begin{subequations}\label{eq:bethe_factorisation}
\begin{align}
 q(s) &= \prod_{a \in \mathcal{V}}q_a(s_a) \prod_{i \in \mathcal{E}}q_i(s_i)^{-1} \\         
 \int q_i(s_i)\mathrm{d}s_i &= 1, \quad \forall i \in \mathcal{E}\\
\int q_a(s_a)\mathrm{d}s_a &=  1, \quad \forall a \in \mathcal{V}  \\
\int q_a(s_a)\mathrm{d}s_{a\backslash i} &=  q_i(s_i), \quad \forall a \in \mathcal{V} ,\, \forall i \in a  
\end{align}
\end{subequations}
on the set $\mathcal{Q}$ \citep{yedidia_bethe_2001}. In \eqref{eq:bethe_factorisation}, $\mathcal{E}$ is a set of variables in a model, $\mathcal{V}$ is a set of local functions $f_a$ in the corresponding TFFG, $(s_a, y_a)$ is a set of (both latent and observed) variables connected to the corresponding node of $f_a$, $q_a(s_a)$ refers to a local variational posterior for factor $f_a(s_a, y_a)$, and $q_i(s_i)$ is a local variational posterior for marginal $p(s_i|y=\hat{y})$.

We refer to the Bethe-constrained variational family $\mathcal{Q}$ as $\mathcal{Q}_B$. The VFE optimisation procedure in \eqref{eq:variational_free_energy_optimisation} with Bethe assumption of \eqref{eq:bethe_factorisation} and extra factorisation and form constraints for the local variational distributions $q_a(s_a)$ and $q_i(s_i)$ is called the \textit{Constrained Bethe Free Energy} (CBFE) optimisation procedure. Many of the well-known message passing algorithms, including Belief Propagation, message passing-based Expectation Maximisation, structured VMP, and others, can be framed as CFBE minimization tasks \citep{yedidia_constructing_2005, zhang_unifying_2017, senoz_variational_2021}. Moreover, different message passing algorithms in an TFFG can straightforwardly be combined, leading to a very large set of possible \emph{hybrid} message passing-based algorithms. All these hybrid variants still allow for a proper and consistent interpretation by performing message passing-based inference through CBFE minimization. Thus, the CBFE minimisation framework provides a principled way to derive new algorithms within message passing-based algorithms and provides a solid foundation for a reactive message passing implementation.

\section{Reactive Message Passing}\label{section:rmp}

This section provides an introduction to reactive programming and describes the core ideas of the reactive message passing framework. First, we discuss the reactive programming approach as a standalone programming paradigm without its relation to Bayesian inference (Section \ref{section:rp_background}). Then we discuss how to connect message passing-based Bayesian Inference with reactive programming (Section \ref{section:rmp_framework}).

\subsection{Reactive Programming}\label{section:rp_background}

This section briefly describes the essential concepts of the Reactive Programming (RP) paradigm such as observables, subscriptions, actors, subjects, and operators. Generally speaking, RP provides a set of guidelines and ideas to simplify working with reactive systems and asynchronous streams of data and/or events. Many modern programming languages have some support for RP through extra libraries and/or packages\footnote{ReactiveX programming languages support \url{http://reactivex.io/languages.html}}. For example, the JavaScript community features a popular RxJS\footnote{JavaScript reactive extensions library \url{https://github.com/ReactiveX/rxjs}} library (more than 26 thousands stars on GitHub) and the Python language community has its own RxPY\footnote{Python reactive extensions library \url{https://github.com/ReactiveX/RxPY}} package. 

\begin{figure}[h]
  \centering
  \hspace*{\fill}%
  \begin{subfigure}[t]{.45\textwidth}
    \includegraphics[width=\columnwidth]{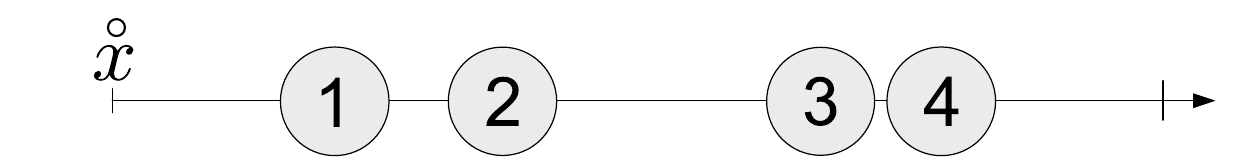}
    \caption{An observable emitting integer values}
    \label{fig:reactive_observable_raw}
  \end{subfigure}
  \hspace*{\fill}%
  \begin{subfigure}[t]{.45\textwidth}
    \includegraphics[width=\columnwidth]{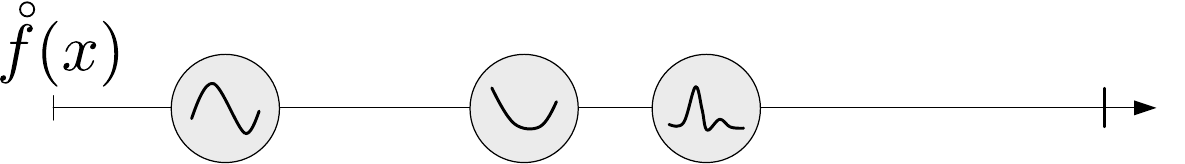}
    \caption{An observable emitting functions of $x$}
    \label{fig:reactive_observable_functions}
  \end{subfigure}
  \hspace*{\fill}%
  \caption{A visual representation of an observable collection over time. An arrow represents a timeline. Circles denote updates at a specific point on that timeline. Values inside circles denote the corresponding data of the update. The bar at the end of the timeline indicates a completion event after which the observable stops sending new updates.}
  \label{fig:reactive_observable_raw_and_functions}
\end{figure}

\paragraph{Observables} The core idea of the reactive programming paradigm is to replace variables in the context of programming languages with observables. Observables are \textit{lazy push collections}, whereas regular data structures such as arrays, lists or dictionaries are \emph{pull} collections. The term pull refers here to the fact that a user may directly ask for the state of these data structures or ``pull'' their values. Observables, in contrast, \textit{push} or \textit{emit} their values over time and it is not possible to directly ask for the state of an observable, but only to \textit{subscribe} to future updates. The term \textit{lazy} means that an observable does not produce any data and does not consume computing resources if no one has subscribed to its updates. Figure~\ref{fig:reactive_observable_raw_and_functions} shows a standard visual representation of observables in the reactive programming context. We denote an observable of some variable $x$ as $\obs{x}$. If an observable emits functions (e.g., a probability density function) rather than just raw values such as floats or integers, we write such observable as $\obs{f}(x)$. This notation does not imply that the observable is a function over $x$, but instead that it emits functions over $x$. 

\begin{figure}[h]
  \centering
  \hspace*{\fill}%
  \begin{subfigure}[t]{.45\textwidth}
    \includegraphics[width=\columnwidth]{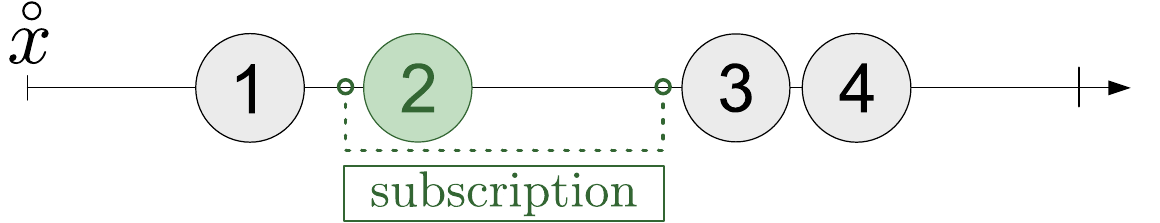}
    \caption{The subscription happens at a specific point in time and allows actors to receive new values. In this example, an actor would receive only update \textcircled{\raisebox{-0.9pt}{2}}, since the subscription was executed after update \textcircled{\raisebox{-0.9pt}{1}} but was cancelled before updates \textcircled{\raisebox{-0.9pt}{3}}, \textcircled{\raisebox{-0.9pt}{4}}.}
    \label{fig:reactive_subscription}
  \end{subfigure}
  \hspace*{\fill}%
  \begin{subfigure}[t]{.45\textwidth}
    \includegraphics[width=\columnwidth]{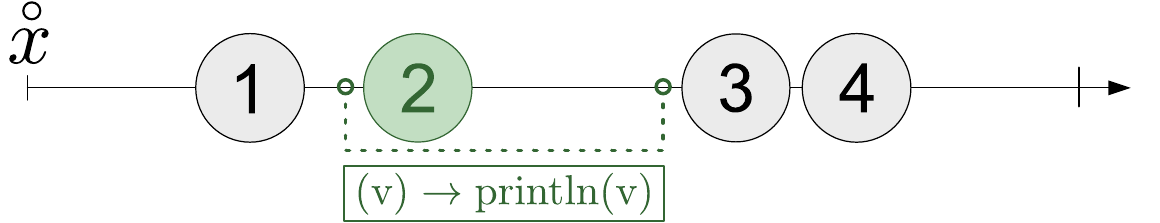}
    \caption{An actor subscribes to an observable, listens to its updates, reacts and performs some actions. In this example the actor simply prints one single value it receives and unsubscribes.}
    \label{fig:reactive_actor}
  \end{subfigure}
  \hspace*{\fill}%
  \caption{A visual representation of a subscription execution and a simple actor. An arrow represents a timeline. Circles denote updates at a specific point on a timeline. Values inside circles denote the corresponding data of the update. The bar at the end of the timeline indicates a completion event after which the observable stops sending new updates.}
  \label{fig:reactive_subscription_and_actor}
\end{figure}

\paragraph{Subscriptions} To start listening to new updates from some observable RP uses subscriptions. A subscription (Fig.~\ref{fig:reactive_subscription}) represents the execution of an observable. Each subscription creates an independent observable execution and consumes computing resources. 

\paragraph{Actors} An actor (Fig.~\ref{fig:reactive_actor}) is a special computational unit that subscribes to an observable and performs some \textit{actions} whenever it receives a new update. In the literature actors are also referred as \textit{subscribers} or \textit{listeners}.

\paragraph{Subjects} A subject is a special type of actor that receives an update and simultaneously re-emits the same update to multiple actors. We refer to those actors as \textit{inner actors}. In other words, a subject utilizes a single subscription to some observable and replicates updates from that observable to multiple subscribers. This process of re-emitting all incoming updates is called \textit{multicasting}. One of the goals of a subject is to share the same observable execution and, therefore, save computer resources. A subject is effectively an actor and observable at the same time.

\paragraph{Operators} An operator is a function that takes a single or multiple observables as its input and returns another observable. An operator is always a pure operation in the sense that the input observables remain unmodified. Next, we discuss a few operator examples that are essential for the reactive message passing framework. 

\begin{figure}[h]
  \centering
  \hspace*{\fill}%
  \begin{subfigure}[t]{.45\textwidth}
    \includegraphics[width=\columnwidth]{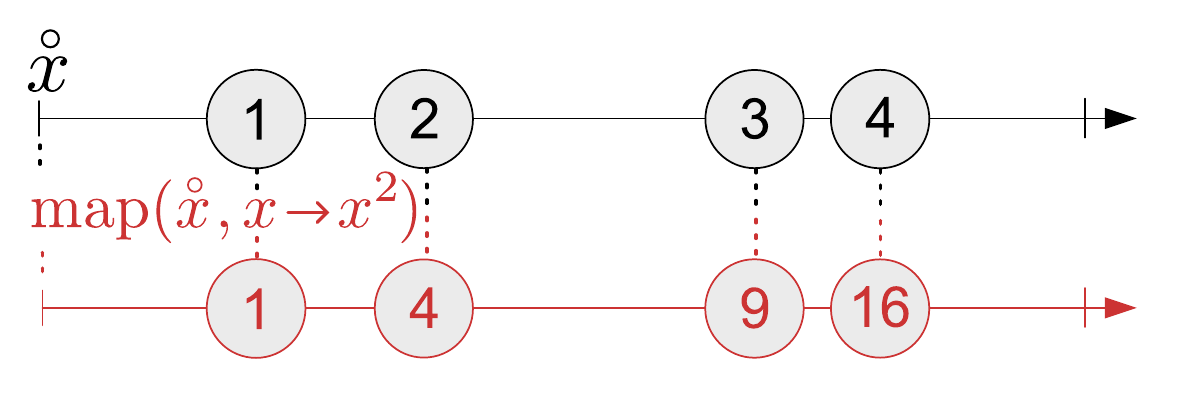}
    \caption{The \texttt{map} operator creates a new observable that mirrors the original observable but with transformed values using the provided mapping function.}
    \label{fig:reactive_operator_map}
  \end{subfigure}
  \hspace*{\fill}%
  \begin{subfigure}[t]{.45\textwidth}
    \includegraphics[width=\columnwidth]{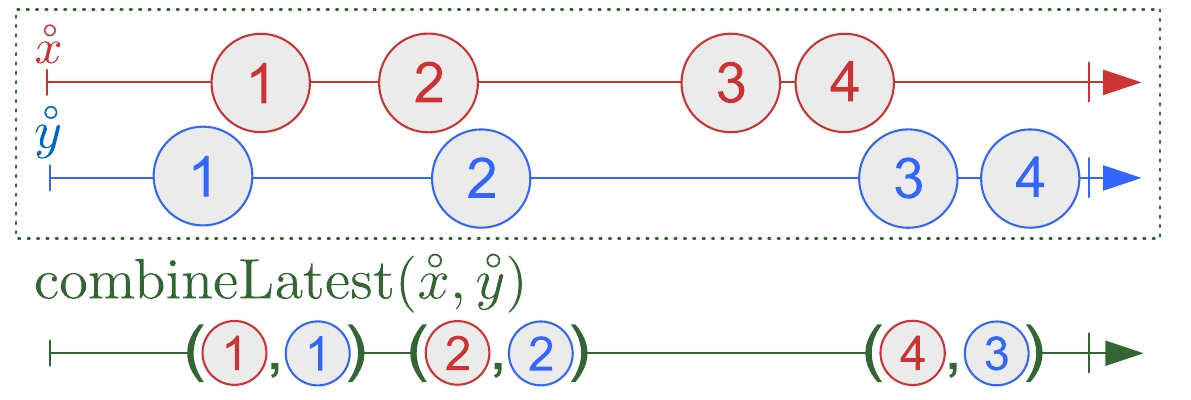}
    \caption{The \texttt{combineLatest} operator combines two or more source observables into a single one and emits a combination of the latest values of all inner source observables.}
    \label{fig:reactive_operator_combine_latest}
  \end{subfigure}
  \hspace*{\fill}%
  \caption{A visual representation of the reactive operators application. All reactive operators do not change the original observables. It is still possible to subscribe to the original observables and observe their values over time.}
  \label{fig:reactive_operators}
\end{figure}

\subparagraph{map} The \texttt{map} (Fig.~\ref{fig:reactive_operator_map}) operator applies a given \textit{mapping function} to each value emitted by a source observable and returns a new modified observable. When we apply the \texttt{map} operator with mapping function $f$ to an observable we say that this observable is \textit{mapped} by a function $f$, see Listing~\ref{lst:map_example}.

\begin{adjustbox}{minipage=\textwidth,margin=0pt \smallskipamount,center}
  \begin{jllisting}[caption={An example of the \texttt{map} operator application. We apply the \texttt{map} operator to obtain a new observable that emits squared values from the original observable. The original observable remains unmodified.}, label={lst:map_example}, captionpos=b]
  # We create the mapped observable of squared values
  # of the original `source` observable
  source         = get_source()
  squared_source = map(source, x -> x^2) 
  \end{jllisting}
\end{adjustbox}

\subparagraph{combineLatest} The \texttt{combineLatest} (Fig.~\ref{fig:reactive_operator_combine_latest}) operator combines multiple observables to create a new observable whose values are calculated from the latest values of each of the original input observables. We refer to those input observables as \textit{inner observables}. We show an example of \texttt{combineLatest} operator application in Listing~\ref{lst:combine_latest_example}. 

\begin{adjustbox}{minipage=\textwidth,margin=0pt \smallskipamount,center}
\begin{jllisting}[caption={An example of the \texttt{combineLatest} operator application. We use the \texttt{combineLatest} operator to combine the latest values from two integer streams and additionally apply the \texttt{map} operator. The resulting observable emits the sum of squared values of $x$ and $y$ as soon as both of them emit a new value.}, label={lst:combine_latest_example}, captionpos=b]
  # We assume that `source1` and `source2` are both observables of integers
  source1 = get_source1()
  source2 = get_source2()

  # We create a new observable by applying a combineLatest operator to it
  combined = combineLatest(source1, source2)

  # We can go further and apply a `map` operator to the combined observable
  # and create a stream of the sum of squares of the latest values 
  # from `source1` and `source2` 
  combined_sum = map(combined, (x1, x2) -> x1^2 + x2^2)
\end{jllisting}
\end{adjustbox}

The reactive programming paradigm has a lot of small and basic operators, e.g., \texttt{filter}, \texttt{count}, \texttt{start\_with}, but the \texttt{map} and the \texttt{combineLatest} operators form the foundation for the RMP framework.

\subsection{The Reactive Message Passing Framework}\label{section:rmp_framework}

In the next section we show how to formulate message passing-based inference in the reactive programming paradigm. First, we cast all variables of interest to observables. These include messages $\mu_{f_j \rightarrow x_i}(x_i)$ and $\mu_{x_i \rightarrow f_j}(x_i)$, marginals $q(s_i)$ over all model variables, and local variational posteriors $q_a(s_a)$ for all factors in the corresponding TFFG of a probabilistic generative model. Next, we show that nodes and edges are, in fact, just special types of subjects that multicast messages with the \texttt{combineLatest} operator and compute corresponding integrals with the \texttt{map} operator. We use the subscription mechanism to start listening for new posterior updates in a model and perform actions based on new observations. The resulting reactive system resolves message (passing) updates locally, automatically reacts on flowing messages and updates itself accordingly when new data arrives. 

\subsubsection{Factor Node Updates}\label{section:factor_node_updates}

In the context of RMP and TFFG, a node consists of a set of connected edges, where each edge refers to a variable in a probabilistic model. The main purpose of a node in the reactive message passing framework is to accumulate updates from all connected edges in the form of message observables $\obs{\mu}_{s_i \rightarrow f}(s_i)$ and marginal observables $\obs{q}(s_i)$ observables with \texttt{combineLatest} operator, followed by computing the corresponding outbound messages with \texttt{map} operator. We refer to a combination of \texttt{combineLatest} and \texttt{map} operators as the \textit{default computational pipeline}.  

\begin{figure}[h]
  \centering
  \hspace*{\fill}%
  \begin{subfigure}[t]{.45\textwidth}
    \includegraphics[width=\columnwidth]{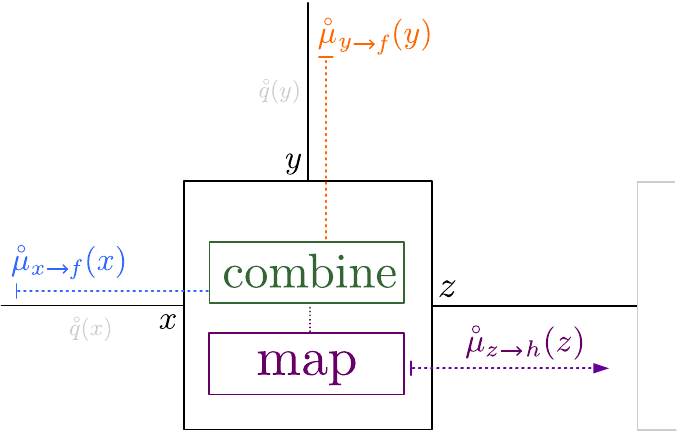}
    \caption{No extra factorisation assumption for local variational distribution $q$ corresponds to the belief propagation update rule \eqref{eq:node_outbound_message_update_bp}.}
    \label{fig:factor_node_outbound_observable_sum_product}
  \end{subfigure}
  \hspace*{\fill}%
  \begin{subfigure}[t]{.45\textwidth}
    \includegraphics[width=\columnwidth]{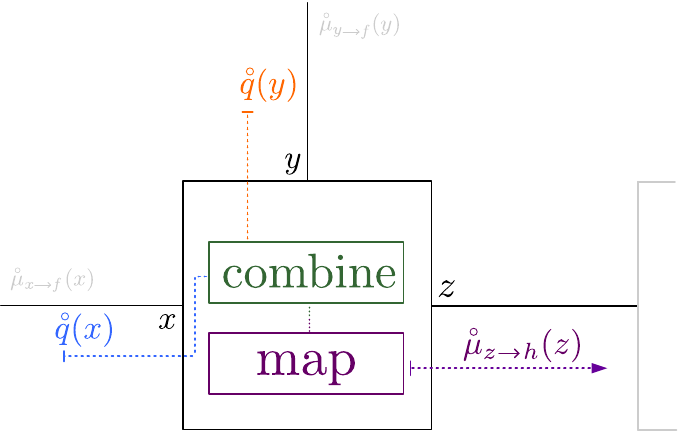}
    \caption{The mean-field factorisation assumption for the local variational distribution $q$ corresponds to the variational message passing update rule \eqref{eq:node_outbound_message_update_vmp}.}
    \label{fig:factor_node_outbound_observable_mean_field}
  \end{subfigure}
  \hspace*{\fill}%
  \caption{Graphical representation of the default computational pipeline for different local factorisation constraints. Different combination of local input observable sources potentially specifies different message passing algorithms.}
  \label{fig:factor_node_outbound_observable}
\end{figure}
 
To support different message passing algorithms, each node handles a special object that we refer to as a \textit{node context}. The node context essentially consists of the local factorisation, the outbound messages form constraints, and optional modifications to the default computational pipeline. In this setting the reactive node decides, depending on the factorisation constraints, \emph{what} to react on and, depending on the form constraints, \emph{how} to react. As a simple example, consider the belief propagation algorithm with the following message computation rule:

\begin{equation}\label{eq:node_outbound_message_update_bp}
  \mu_{z \rightarrow h}(z) = \int \mu_{x \rightarrow f}(x)\mu_{y \rightarrow f}(y)f(x, y, z)\mathrm{d}x\mathrm{d}y\,,
\end{equation}

where $f$ is a functional form of a node with three arguments $x, y, z$, $\mu_{z \rightarrow h}(z)$ is an outbound message on edge $z$ towards node $h$, and $\mu_{x \rightarrow f}(x)$ and $\mu_{y \rightarrow f}(y)$ are inbound messages on edges $x$ and $y$ respectively. This equation does not depend on the marginals $q(x)$ and $q(y)$ on the corresponding edges, hence, \texttt{combineLatest} needs to take into account only the inbound message observables (Fig.~\ref{fig:factor_node_outbound_observable_sum_product}). In contrast, the variational outbound message update rule for the same node with mean-field factorisation assumption 
\begin{equation}
  \mu_{z \rightarrow h}(z) = \exp \int q(x)q(y)\log f(x, y, z) \mathrm{d}x\mathrm{d}y\,,
  \label{eq:node_outbound_message_update_vmp}
\end{equation}
does not depend on the inbound messages but rather depends only on the corresponding marginals (Fig.~\ref{fig:factor_node_outbound_observable_mean_field}). It is possible to adjust the node context and customize the input arguments of the \texttt{combineLatest} operator such that the outbound message on some particular edge will depend on any subset of local inbound messages, local marginals, and local joint marginals over subset of connected variables. From a theoretical point of view, each distinct combination potentially leads to a new type of message passing algorithm \citep{senoz_variational_2021} and may have different performance characteristics that are beyond the scope of this paper. The main idea is to support as many message passing-based algorithms as possible by employing simple reactive dependencies between local observables.

\begin{jllisting}[label={lst:outbound-message},caption={Pseudo-code for generating an outbound message for each edge connected to a factor node.},captionpos=b,float,floatplacement=H]
  context = getcontext(factornode)
  for edge in local_edges(factornode)
    # Combine all updates from a node's local edges,
    # excluding updates from the current edge
    updates  = combineLatest(local_edges(factornode, context))
    # Create an outbound message observable by applying
    # a map operator with a suitable `compute_message` procedure
    outbound = map(updates, u -> compute_message(u, context))
    ...
  end
\end{jllisting}

\begin{figure}[h]
  \centering
  \hspace*{\fill}%
  \begin{subfigure}{.40\textwidth}
    \includegraphics[width=\columnwidth]{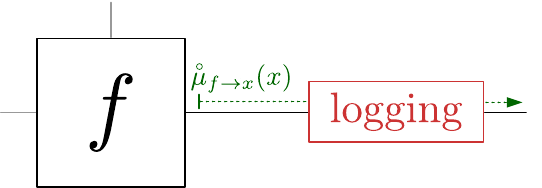}
    \caption{An example of applying a logging operator to an observable of outbound messages.}
    \label{fig:pipeline_logging}
  \end{subfigure}
  \hspace*{\fill}%
  \begin{subfigure}{.40\textwidth}
    \includegraphics[width=\columnwidth]{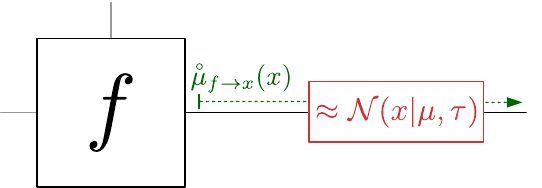}
    \caption{An example of applying a Laplace approximation operator to an observable of outbound messages.}
    \label{fig:pipeline_N}
  \end{subfigure}
  \hspace*{\fill}%
  \caption{A graphical representation of different custom default pipeline modifications for some outbound message observables.}
  \label{fig:pipeline}
\end{figure}

After the local context object is set, a node reacts autonomously on new message or marginal updates and computes the corresponding outbound messages for each connected edge independently, see Listing~\ref{lst:outbound-message}. In addition, outbound message observables naturally support extra operators before or after the \texttt{map} operator. Examples of these custom operators include logging (Fig.~\ref{fig:pipeline_logging}), message approximations (Fig.~\ref{fig:pipeline_N}), and vague message initialisation. Adding extra steps to the default message computational pipeline can help customize the corresponding inference algorithm to achieve even better performance for some custom models \citep{zhang_unifying_2017, akbayrak_reparameterization_2019}. For example, in cases where no exact analytical message update rule exists for some particular message, it is possible to resort to the Laplace approximation in the form of an extra reactive operator for an outbound message observable, see Listing~\ref{lst:custom-pipeline}. Another use case of the pipeline modification is to relax some of the marginalisation constraints from \eqref{eq:bethe_factorisation} with the moment matching constraints
\begin{subequations}
  \begin{align}
  \mathrm{Mean}\left[\int q_a(s_a)\mathrm{d}s_{a\backslash i}\right] & = \mathrm{Mean}\left[q_i(s_i)\right] \\
  \mathrm{Cov}\left[\int q_a(s_a)\mathrm{d}s_{a\backslash i}\right] & = \mathrm{Cov}\left[q_i(s_i)\right],
  \end{align}
\end{subequations}
which are used in the expectation propagation \citep{raymond_expectation_2014, cox_robust_2018} algorithm and usually are easier to satisfy.

\begin{adjustbox}{minipage=\textwidth,margin=0pt \smallskipamount,center}
   \begin{jllisting}[label={lst:custom-pipeline},caption={Pseudo-code for customizing default outbound messages computational pipeline with custom pipeline stages.},captionpos=b]
  context = getcontext(factornode)
  for edge in local_edges(factornode)
    updates  = combineLatest(local_dependecies(factornode, context))
    outbound = updates |> map(u -> compute_message(u, context))
    # Here we apply a custom approximation operator 
    outbound = outbound |> approximate_as_gaussian()
    # or logging statements
    outbound = outbound |> log_into_file()
    # or moment matching constraint
    outbound = outbound |> moment_matching()
    ...
  end
  \end{jllisting}
\end{adjustbox}

\subsubsection{Marginal Updates}\label{section:marginal_updates}

\begin{figure}[h]
  \centering
  \includegraphics[width=0.65\columnwidth]{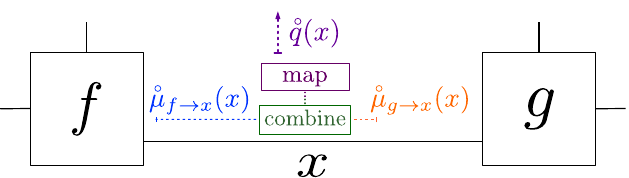}
  \caption{A visual representation of marginal observable computation for some variable $x$ and its corresponding edge with the default computational pipeline. The marginal distribution for a variable is equal to the product of two colliding messages on the corresponding edge divided by a normalisation constant.}
  \label{fig:marginals_single_edge}
\end{figure}

In the context of RMP and TTFG, each edge refers to a single variable $s_i$, reacts to outbound messages from connected nodes, and emits the corresponding marginal $q(s_i)$. Therefore, we define the marginal observable $\obs{q}(s_i)$ as a combination of two adjacent outbound message observables from connected nodes with the \texttt{combineLatest} operator and their corresponding normalised product with the \texttt{map} operator, see Listing~\ref{lst:edge-posterior} and Fig.~\ref{fig:marginals_single_edge}. Regarding node updates, we refer to the combination of the \texttt{combineLatest} and \texttt{map} operators as the default computational pipeline. Each edge has its own \textit{edge context} object that is used to customise the default pipeline. Extra pipeline stages are mostly needed to assign extra form constraints for variables in a probabilistic model as discussed in Section~\ref{section:background}. For example, in the case where no analytical closed-form solution is available for a normalized product of two colliding messages, it is possible to modify the default computational pipeline and to fall back to an available approximation method, such as the point-mass form constraint that leads to the Expectation Maximisation algorithm, or Laplace Approximation.

\begin{adjustbox}{minipage=\textwidth,margin=0pt \smallskipamount,center}
  \begin{jllisting}[label={lst:edge-posterior},caption={Pseudo-code for generating an posterior marginal update for each edge in a probabilistic model.},captionpos=b]
    for edge in edges(model)
      context = getcontext(edge)

      left_message  = get_left_message_updates(edge, context)  
      right_message = get_right_message_updates(edge, context)

      # Combine all updates from a left and right observables
      updates = combineLatest(left_message, right_message)

      # a map operator with a `prod_and_normalise` procedure
      posterior = map(updates, u -> prod_and_normalise(u, context))
    end
  \end{jllisting}
\end{adjustbox}

\subsubsection{Joint Marginal Updates}

\begin{figure}[h]
  \centering
  \includegraphics[width=0.65\columnwidth]{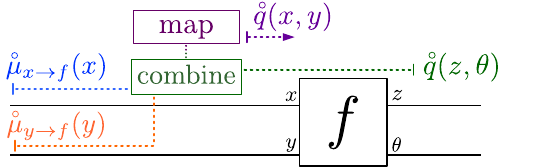}
  \caption{Visual representation of the local variational distribution observable computation on a subset of edges around a node $f$ with four edges $x$, $y$, $z$, and $\theta$. The local variational distribution for a subset of connected edges depends on inbound messages on these edges and the local variational distribution over remaining edges and corresponding variables.}
  \label{fig:marginals_joint}
\end{figure}

The procedure for computing a local marginal distribution for a subset of connected variables around a node in TFFG is similar to previous cases and has the \texttt{combineLatest} and \texttt{map} operators in its default computational pipeline. To compute a local marginal distribution $q(s_{a\setminus b})$, we need local inbound messages updates to the node $f$ from edges in $s_{a \setminus b}$ and local joint marginal $q(s_b)$. In Fig.~\ref{fig:marginals_joint}, we show an example of a local joint marginal observable where $s_a = \{x,y,z,\theta\}$, $s_{a \setminus b} = \{x, y\}$, and $s_b = \{z, \theta\}$. 

\subsubsection{Bethe Free Energy Updates}

The BFE is usually a good approximation to the VFE, which is an upper bound to Bayesian model evidence, which in turn scores the performance of a model. Therefore, it is often useful to compute BFE. To do so, we first decompose \eqref{eq:VFE} into a sum of node local energy terms $U[f_a, q_a]$ and sum of local variable entropy terms $H[q_i]$:

\begin{subequations}
    \begin{align}
    U[f_a, q_a] & = \int q_a(s_a)\log \frac{q_a(s_a)}{f_a(s_a)}\mathrm{d}s_a,\\
    H[q_i] & = \int q_i(s_i)\log \frac{1}{q_i(s_i)}\mathrm{d}s_i,\\
    F[q] & = \sum_{a \in \mathcal{V}} U[f_a, q_a] + \sum_{i \in \mathcal{E}} H[q_i].\label{eq:bfe_updates_subeq_decomposed}
    \end{align}
\end{subequations}

\begin{figure}[h]
    \centering
    \includegraphics[width=0.95\columnwidth]{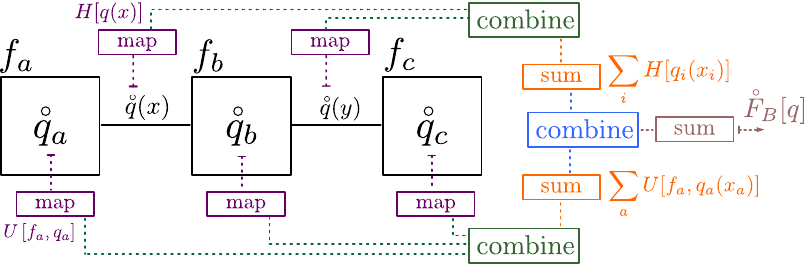}
    \caption{A visual representation of the Bethe Free Energy observable computation for an arbitrary TFFG with nodes $f_a$, $f_b$ and $f_c$, together with random variables $x$ and $y$. For each node, the algorithm reacts on observables of variational distributions $\mathring{q}_a$, $\mathring{q}_b$ and $\mathring{q}_c$ and computes corresponding average energy terms with the \texttt{map} operator. For each edge, the algorithm reacts on observables of marginal distributions $\mathring{q}(x)$ and $\mathring{q}(y)$ and computes corresponding entropy terms with the \texttt{map} operator. The resulting updates are combined and transformed with the \texttt{combineLatest} and the \texttt{sum} operators, respectively.}
    \label{fig:free_energy}
\end{figure}

We then use the \texttt{combineLatest} operator to combine the local joint marginal observables and marginal observables for each individual variable in a probabilistic model, followed by using the \texttt{map} operator to compute the average energy and entropy terms, see Fig.~\ref{fig:free_energy} and Listing~\ref{lst:bfe-observable}. Each combination emits an array of numbers as an update and we make usage of the \texttt{sum} reactive operator that computes the sum of such updates.

\begin{adjustbox}{minipage=\textwidth,margin=0pt \smallskipamount,center}
\begin{jllisting}[label={lst:bfe-observable}, language=julia, style=jlcodestyle, caption={Pseudo-code for creating the BFE observable \eqref{eq:bfe_updates_subeq_decomposed} for an arbitrary TFFG.},captionpos=b]
  # U_local is an array of node local average energies observables
  U_local = foreach(factor_nodes) do f_a
      return map(local_q(f_a), (q_a) -> compute_energy(f_a, q_a))
  end
  
  # H_local is an array of edge local entropies observables
  H_local = foreach(variables) do variable
      return map(local_q(variable), (q_i) -> compute_entropy(q_i))
  end
  
  # `sum` returns an observable 
  U = sum(combineLatest(U_local)) # Total average energy observable 
  H = sum(combineLatest(H_local)) # Total entropies observable 
  
  # The resulting BFE observable emits a sum of 
  # node local average energies and edge local entropies
  bfe = sum(combineLatest(U, H))
\end{jllisting}
\end{adjustbox}

This procedure assumes that we are able to compute average energy terms for any local joint marginal around any node and to compute entropies of all resulting variable marginals in a probabilistic model. The resulting BFE observable from Listing~\ref{lst:bfe-observable} autonomously reacts on new updates in the probabilistic model and emits a new value as soon as we have new updates for local variational distributions $q_a(s_a)$ and local marginals $q_i(s_i)$.

\subsubsection{Infinite Reaction Chain Processing}\label{section:infinite_reaction_chain_processing}

Procedures from the previous sections create observables that may depend on itself, especially in the case if an TFFG has a loop in it. This potentially creates an infinite reaction chain. However, the RMP framework naturally handles such cases and infinite data stream processing in its core design as it uses reactive programming as its foundation. In general, the RP does not make any assumptions on the underlying nature of the update generating process of observables. Furthermore, it is possible to create a recursive chain of observables where new updates in one observable depend on updates in another or even on updates in the same observable. Any reactive programming implementation in any programming language supports infinite chain reaction by design and still provides capabilities to control such a reactive system. 

Furthermore, in view of graphical probabilistic models for signal processing applications, we may create a single time section of a Markov model and redirect a posterior update observable to a prior observable, effectively creating an infinite reaction chain that mimics an infinite factor graph with messages flowing only in a forward-time direction. This mechanism makes it possible to use infinite data streams and to perform Bayesian inference in real-time on infinite graphs as soon as the data arrives. We will show an example of such an inference application in the Section~\ref{section:experiment_hgf}. 

\begin{figure}[h]
  \centering
  \includegraphics[width=0.66\columnwidth]{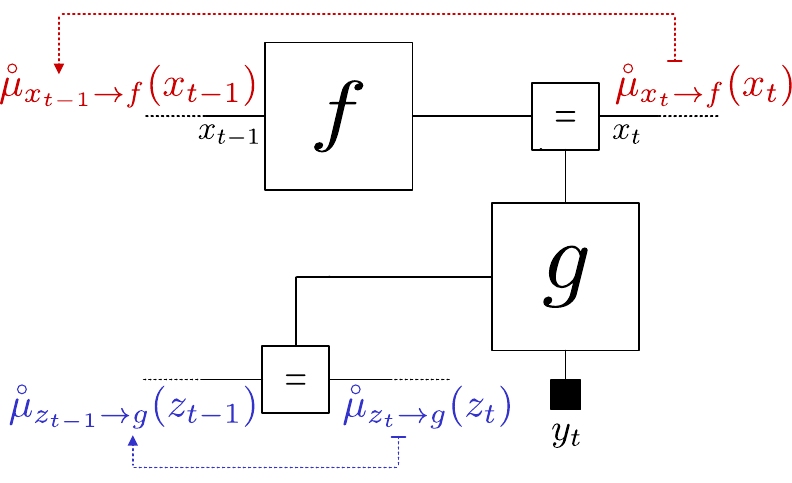}
  \caption{An example of an infinite factor graph in the RMP framework with arbitrary state transition nodes $f$ and $g$. The small black square indicates observed variable. The dotted red line indicates a redirection of an observable of posterior updates for $x_{t}$ to an observable of priors for $x_{t - 1}$. The dotted red line indicates a redirection of an observable of posterior updates for $z_{t}$ to an observable of priors for $z_{t - 1}$. The model reacts on new incoming data points $y_t$ and computes messages in time-forward direction only.}
  \label{fig:graph_inifinite}
\end{figure}

This setting creates an infinite stream of posteriors redirected to a stream of priors. The resulting system reacts on new observations $y_t$ and computes messages automatically as soon as a new data point is available. However, in between new observations, the system stays idle and simply waits. In case of variational Bayesian inference, we may choose to perform more VMP iterations during idle time so as to improve the approximation of \eqref{eq:marginal_posterior_distibutions}. 

\section{Implementation}\label{section:implementation}

Based on the Sections \ref{section:motivation} and \ref{section:background}, a generic toolbox for message passing-based Constrained Bethe Free Energy minimization on TFFG representations of probabilistic models should at least comprise the following features:
\begin{enumerate}
    \itemsep0em
    \item A comprehensive specification language for probabilistic models $p(y, s)$ of interest;
    \item A convenient way to specify additional local constraints on $\mathcal{Q}_B$ so as to restrict the search space for posterior $q(s)$. Technically, items 1 and 2 together support specification of a CBFE optimisation procedure;
    \item Provide an automated, efficient and scalable engine to minimize the CBFE by message passing-based inference.
\end{enumerate}

Implementation of all three features in a user-friendly and efficient manner is a very challenging problem. We have developed a new reactive message passing-based Bayesian inference package, which is fully written in the open-source, high-performant scientific programming language Julia \citep{bezanson_julia:_2017}. Our implementation makes use of the Rocket.jl package, which we developed to support the reactive programming in Julia. The Rocket.jl implements observables, subscriptions, actors, basic reactive operators such as the \texttt{combineLatest} and the \texttt{map}, and makes it easier to work with asynchronous data streams. The framework exploits the Distributions.jl \citep{besancon_distributionsjl_2021, lin_juliastatsdistributionsjl_2019} package that implements various probability distributions in Julia. We also use the GraphPPL.jl package to simplify the probabilistic model specification and variational family constraints specification parts. We refer to this ecosystem as a whole as \textbf{ReactiveMP.jl}\footnote{ReactiveMP.jl GitHub repository \url{https://github.com/biaslab/ReactiveMP.jl}}.

\subsection{Model specification}

ReactiveMP.jl uses a \texttt{@model} macro for joint model and variational family constraints specification. The \texttt{@model} macro returns a function that generates the model with specified constraints. It has been designed to resemble existing probabilistic model specification libraries like Turing.jl closely, but also to be flexible enough to support reactive message passing-based inference as well as factorisation and form constraints for variational families $\mathcal{Q}_B$ from \eqref{eq:bethe_factorisation}. As an example of the model specification syntax, consider the simple example in Listing~\ref{lst:graphppl_first_example}:

\begin{adjustbox}{minipage=\textwidth,margin=0pt \smallskipamount,center}
\begin{jllisting}[label={lst:graphppl_first_example}]
# We use the `@model` macro to accept a Julia function as input.
# In this example, the model accepts an additional argument `n` that denotes
# the number of observations in the dataset
@model function coin_model(n)
  # Reactive data inputs for Beta prior over θ random variable
  # so we don't need actual data at model creation time
  a = datavar(Float64)
  b = datavar(Float64)
  
  # We use the tilde operator to define a probabilistic relationship between
  # random variables and data inputs. It automatically creates a new random 
  # variable in the current model and the corresponding factor nodes
  θ ~ Beta(a, b) # creates both `θ` random variable and Beta factor node

  # A sequence of observations with length `n`
  y = datavar(Float64, n)

  # Each observation is modelled by a `Bernoulli` distribution 
  # that is governed by a `θ` parameter
  for i in 1:n
    y[i] ~ Bernoulli(θ)  # Reuses `y[i]` and `θ` and creates `Bernoulli` node
  end
  
  # `@model` function must have a `return` statement.
  # Later on, the returned references might be useful to obtain the marginal
  # updates over variables in the model and to pass new observations to data 
  # inputs so the model can react to changes in data
  return y, a, b, θ
end

# The `@model` macro modifies the original function to return a reference 
# to the actual graphical probabilistic model as a factor graph 
# and the exactly same output as in the original `return` statement 
model, (y, a, b, θ) = coin_model(100)
\end{jllisting}
\end{adjustbox}

The tilde operator ($\sim$) creates a new random variable with a given functional relationship with other variables in a model and can be read as "is sampled from ..." or "is modelled by ...". Sometimes it is useful to create a random variable in advance or to create a sequence of random variables in a probabilistic model for improved efficiency. The ReactiveMP.jl package exports a \texttt{randomvar()} function to manually create random variables in the model.

\begin{adjustbox}{minipage=\textwidth,margin=0pt \smallskipamount,center}
\begin{jllisting}
x = randomvar()     # A single random variable in the model
x = randomvar(n)    # A collection of n random variables  
\end{jllisting}
\end{adjustbox}

To create observations in a model, the ReactiveMP.jl package exports the \texttt{datavar(T)} function, where \texttt{T} refers to a type of data such as \texttt{Float64} or \texttt{Vector\{Float64\}}. Later on it is important to continuously update the model with new observations to perform reactive message passing.

\begin{adjustbox}{minipage=\textwidth,margin=0pt \smallskipamount,center}
\begin{jllisting}
y = datavar(Float64)         # A single data input of type Float64
y = datavar(Vector{Float64}) # A single data input of type Vector{Float64}
y = datavar(Float64, n)      # A sequence of n data inputs of type Float64
\end{jllisting}
\end{adjustbox}

The tilde operator supports an optional \jlinl{where} statement to define local node- or variable-context objects. As discussed in Section \ref{section:rmp_framework}, local context objects specify local factorisation constraints for the variational family of distributions $\mathcal{Q}_B$, or may include extra pipeline stages to the default computational pipeline of messages and marginals updates, see Listing~\ref{lst:graphppl_where_example}. 

\begin{adjustbox}{minipage=\textwidth,margin=0pt \smallskipamount,center}
\begin{jllisting}[label={lst:graphppl_where_example}]
# We can use `where` clause to specify extra factorisation
# constraints for the local variational distribution
θ ~ Beta(a, b) where { q = q(θ)q(a)q(b) }

# Structured factorisation assumption 
x_next ~ NormalMeanVariance(x_previous, tau) where { 
  q = q(x_next, x_previous)q(tau) 
}

# We can also use `where` clause to modify the default 
# computational pipeline for all outbound messages from a node
# In this example `LoggerPipelineStage` modifies pipeline such 
# that it starts to print all outbound messages into standard output
y ~ Gamma(a, b) where { pipeline = LoggerPipelineStage() }
\end{jllisting}
\end{adjustbox}

\subsection{Marginal updates}\label{section:posterior_updates}

The ReactiveMP.jl framework API exports the \texttt{getmarginal()} function, which returns a reference for an observable of marginal updates. We use the \texttt{subscribe!()} function to subscribe to future marginal updates and perform some actions.

\begin{adjustbox}{minipage=\textwidth,margin=0pt \smallskipamount,center}
\begin{jllisting}
model, (y, a, b, θ) = coin_model()

# `subscribe!` accepts an observable as its first argument and  
# a callback function as its second argument
θ_subscription = subscribe!(getmarginal(θ), (update) -> println(update))
\end{jllisting}
\end{adjustbox}

For variational Bayesian inference and loopy belief propagation algorithms, it might be important to initialise marginal distributions over the model's variables and to initialise messages on edges. The framework exports the \texttt{setmarginal!()} and \texttt{setmessage!()} functions to set initial values to the marginal update observables and the message update observables, respectively.

\begin{adjustbox}{minipage=\textwidth,margin=0pt \smallskipamount,center}
\begin{jllisting}
setmarginal!(θ, Bernoulli(0.5)) 
setmessage!(θ, Bernoulli(0.5))
\end{jllisting}
\end{adjustbox}

\subsection{Bethe Free Energy updates}

The ReactiveMP.jl framework API exports the \texttt{score()} function to get a stream of Bethe Free Energy values. By default, the Bethe Free Energy observable emits a new update only after all variational posterior distributions $q_a(s_a)$ and $q_i(s_i)$ have been updated.

\begin{adjustbox}{minipage=\textwidth,margin=0pt \smallskipamount,center}
\begin{jllisting}
bfe_updates      = score(BetheFreeEnergy(), model)
bfe_subscription = subscribe!(bfe_updates, (update) -> println(update))
\end{jllisting}
\end{adjustbox}

\subsection{Multiple dispatch}

An important problem is how to select the most efficient update rule for an outbound message, given the types of inbound messages and/or marginals. In general, we want to use known analytical solutions if they are available and otherwise select an appropriate approximation method. In the context of software development, the choice of which method to execute when a function is applied is called \textit{dispatch}.

Locality is a central design choice of the reactive message passing approach, but it makes it impossible to infer the correct inbound message types locally before the data have been seen.  Fortunately, the Julia language supports \textit{dynamic multiple dispatch}, which enables an elegant solution for this problem. Julia allows the dispatch process to choose which of a function's methods to call based on the number of arguments given, and on the types of all of the function's arguments at run-time. This feature enables automatic dispatch of the most suitable outbound message computation rule, given the functional form of a factor node and the functional forms of inbound messages and/or posterior marginals.

Julia's built-in features also support the ReactiveMP.jl implementation to dynamically dispatch to the most efficient message update rule for both exact or approximate variational algorithms. If no closed-form analytical message update rule exists, Julia's multiple dispatch facility provides several options to select an alternative (more computationally demanding) update as discussed in the Section \ref{section:factor_node_updates}. 

The ReactiveMP.jl uses the following arguments to dispatch to the most efficient message update rule (see Listing.~\ref{code:rule_md}):
\begin{itemize}
  \itemsep0em
  \item Functional form of a factor node: e.g. \texttt{Gaussian} or \texttt{Gamma};
  \item Name of the connected edge: e.g. \texttt{:out};
  \item Local variable constraints: e.g. \texttt{Marginalisation} or \texttt{MomentMatching};
  \item Input messages names and types with \texttt{m\_} prefix: e.g. \texttt{m\_mean::Gaussian};
  \item Input marginals names and types with \texttt{q\_} prefix: e.g. \texttt{q\_precision::Any};
  \item Optional context object, e.g. a strategy to correct nonpositive definite matrices, an order of autoregressive model or an optional approximation methods used to compute messages.
\end{itemize}

\begin{jllisting}[label={code:rule_md}, language=julia, style=jlcodestyle, caption={An example of node and VMP message update rule specification for a univariate Gaussian distribution with mean-precision parametrisation.},captionpos=b,float,floatplacement=H] 
# First we need to define a Gaussian node with 3 edges
# We assume mean-precision parametrisation here
@node Gaussian Stochastic [ out, mean, precision ]

# Structured VMP message update rule
@rule Gaussian(:out, Marginalisation) (m_mean::Gaussian, q_precision::Any) = 
begin
  return Gaussian(mean(m_mean), 1 / (var(m_mean) + 1 / mean(q_precision)))
end
\end{jllisting}

All of this together allows the ReactiveMP.jl framework to automatically select the most efficient message update rule. It should be possible to emulate this behavior in other programming languages, but Julia's core support for efficient dynamic multiple dispatch is an important reason why we selected it as the implementation language. It is worth noting that the Julia programming language generates efficient and scalable code with similar run-time performance as for C and C++.

\section{Experimental Evaluation}\label{section:experiments}

In this section, we provide the experimental results of our RMP implementation on various Bayesian inference problems that are common in signal processing applications. The main purpose of this section is to show that reactive message passing and its ReactiveMP.jl implementation in particular is capable of running inference for different common signal processing applications and to explore its performance characteristics.

Each example in this section is self-contained and is aimed to model particular properties of the underlying data sets. The linear Gaussian state space model example in Section \ref{section:experiment_lgssm} models a signal with continuously valued latent states and uses a large static data set with hundreds of thousands of observations. In Section \ref{section:experiment_hmm}, we present a Hidden Markov Model for a discretely valued signal with unknown state transition and observation matrices. The Hierarchical Gaussian Filter example in Section \ref{section:experiment_hgf} shows online learning in a hierarchical model with non-conjugate relationships between variables and makes use of a custom factor node with custom approximate message computation rules. 

For verification, we used synthetically generated data, but the ReactiveMP.jl has been battle-tested on more sophisticated models with real-world data sets \citep{podusenko_message_2021, van_erp_bayesian_2021, podusenko_aida_2022}. Each example in this section has a comprehensive, interactive, and reproducible demo on GitHub\footnote{All experiments are available at \url{https://github.com/biaslab/ReactiveMPPaperExperiments}} experiments repository. More models, tutorials, and advanced usage examples are available in the ReactiveMP.jl package repository on GitHub\footnote{More models, tutorials, and advanced usage examples are available at \url{https://github.com/biaslab/ReactiveMP.jl/tree/master/demo}}.

For each experiment, we compared the performance of the new reactive message passing-based inference engine with another message passing package ForneyLab.jl \citep{van_de_laar_forneylab.jl:_2018} and the sampling-based inference engine Turing.jl \citep{ge_turing_2018}. We selected Turing.jl as a flexible, mature, and convenient platform to run sampling-based inference algorithms in the Julia programming language. In particular, it provides a particle Gibbs sampler for discrete parameters as well as a Hamiltonian Monte Carlo sampler for continuous parameters. We show that the new reactive message passing-based solution not only scales better, but also yields more accurate posterior estimates for the hidden states of the outlined models in comparison with sampling-based methods. Note, however, that Turing.jl is a general-purpose probabilistic programming toolbox and provides instruments to run Bayesian inference in a broader class of probabilistic models than the current implementation of ReactiveMP.jl.

Variational inference algorithms use BFE to evaluate the model's accuracy performance. However, some packages only compute posterior distributions and ignore the BFE. To compare the posterior results for the different Bayesian inference methods that do not compute the BFE functional, we performed the posterior estimation accuracy test by the following metric:

\begin{equation}\label{eq:average_mse}
    AE[q] = \frac{1}{|\mathcal{D}|}\sum_{d \in \mathcal{D}} \left[\frac{1}{T}\sum_{t = 1}^{T} \mathbb{E}_{q(x_t)}[f(x_t - r_t)]\right],
\end{equation}
where $\mathcal{D}$ is a set of all synthetic datasets $d$ for a particular model, $T$ is a number of time steps used in an experiment, $q(x_t)$ is a resulting posterior $p(x_t|y=\hat{y})$ at time step $t$, $r_t$ is an actual value of real underlying signal at time step $t$, $f$ is any positive definite transform. In our experiments, we used $f(x) = x'x$ for continuous multivariate variables, $f(x) = x^2$ for continuous univariate variables and $f(z) = |z|$ for discrete variables. We call this metric an \textit{average error} (AE). We tuned hyperparameters for sampling-based methods in Turing.jl such that to show comparable accuracy results to message passing-based methods in terms of the AE accuracy metric \eqref{eq:average_mse}.

All benchmarks have been performed with help of the BenchmarkTools.jl package \citep{chen_robust_2016}, which is a framework for writing, running, and comparing groups of benchmarks in Julia. For benchmarks, we used current Julia LTS (long-term support) version 1.6 running on a single CPU of a MacBookPro-2018 with a 2.6 GHz Intel Core-i7 and 16GB 2400 MHz DDR4 of RAM. 

\subsection{Linear Gaussian State Space Model}\label{section:experiment_lgssm}

As our first example, we consider the linear Gaussian state space model (LG-SSM) that is widely used in the signal processing and control communities, see \citep{sarkka_bayesian_2013}. The simple LG-SMM is specified by 
\begin{equation} \label{eq:linear_gaussian_ssm}
  \begin{aligned}
    p(x_t|x_{t-1}) &= \mathcal{N}(x_t | Ax_{t-1}, P) \\
    p(y_t|x_{t}) &= \mathcal{N}(y_t | Bx_{t}, Q)   
  \end{aligned}
\end{equation}
where $y_t$ is the observation at time step $t$, $x_t$ represents the latent state at time step $t$, $A$ and $B$ are state transition and observation model matrices, respectively, and $P$ and $Q$ are Gaussian noise covariance matrices. 

\paragraph{Model specification}

\begin{figure}[h]
  \centering
  \hspace*{\fill}%
  \begin{subfigure}[t]{.45\textwidth}
    \includegraphics[width=\columnwidth]{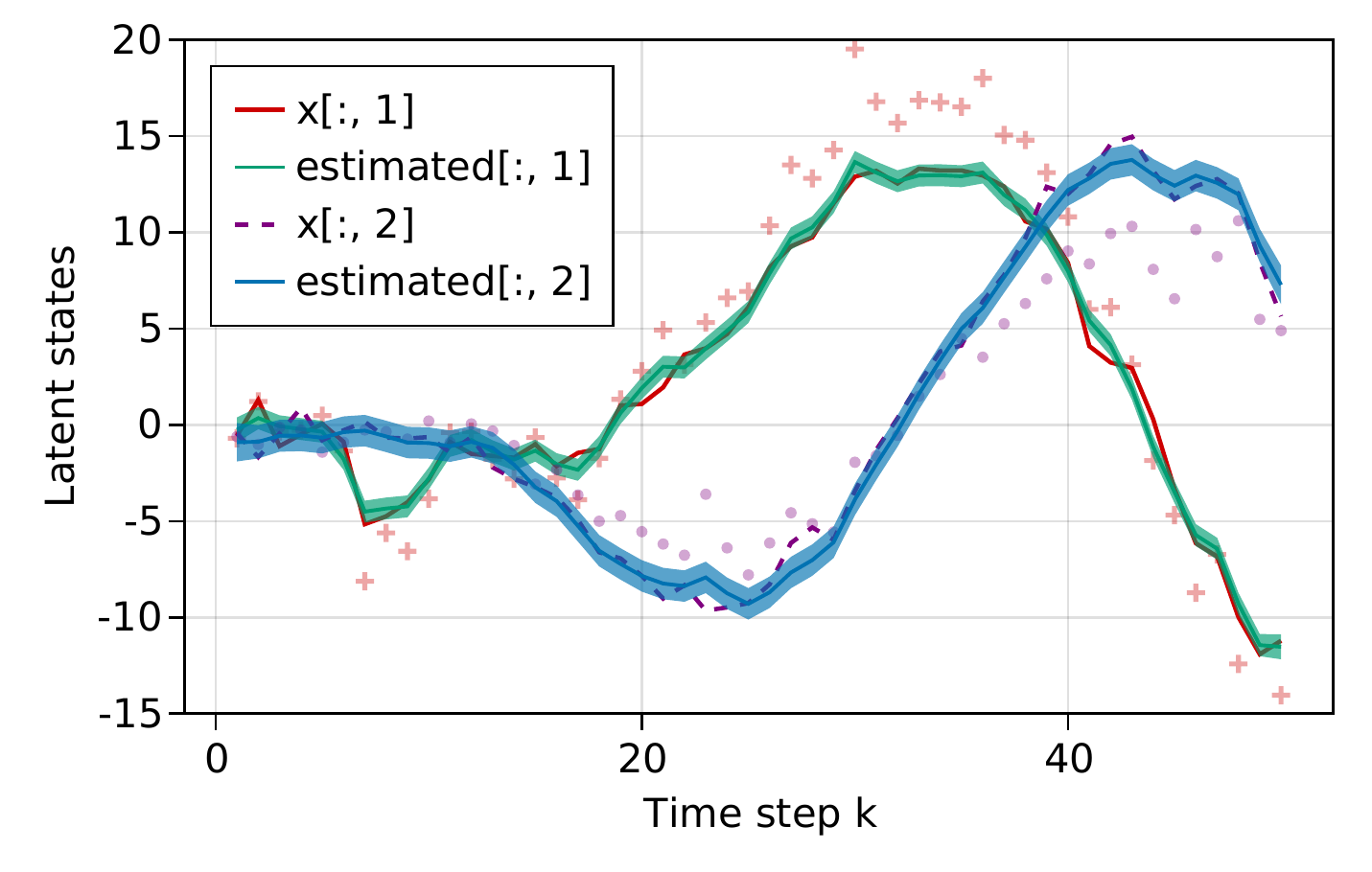}
    \caption{Simulated trajectory, observations and the inference results. The x-axis represents the time steps and the y-axis corresponds to the actual value of hidden states and observations for each dimension.}
    \label{fig:linear_gaussian_ssm_50_smoothing_inference}
  \end{subfigure}
  \hspace*{\fill}%
  \begin{subfigure}[t]{.45\textwidth}
    \includegraphics[width=\columnwidth]{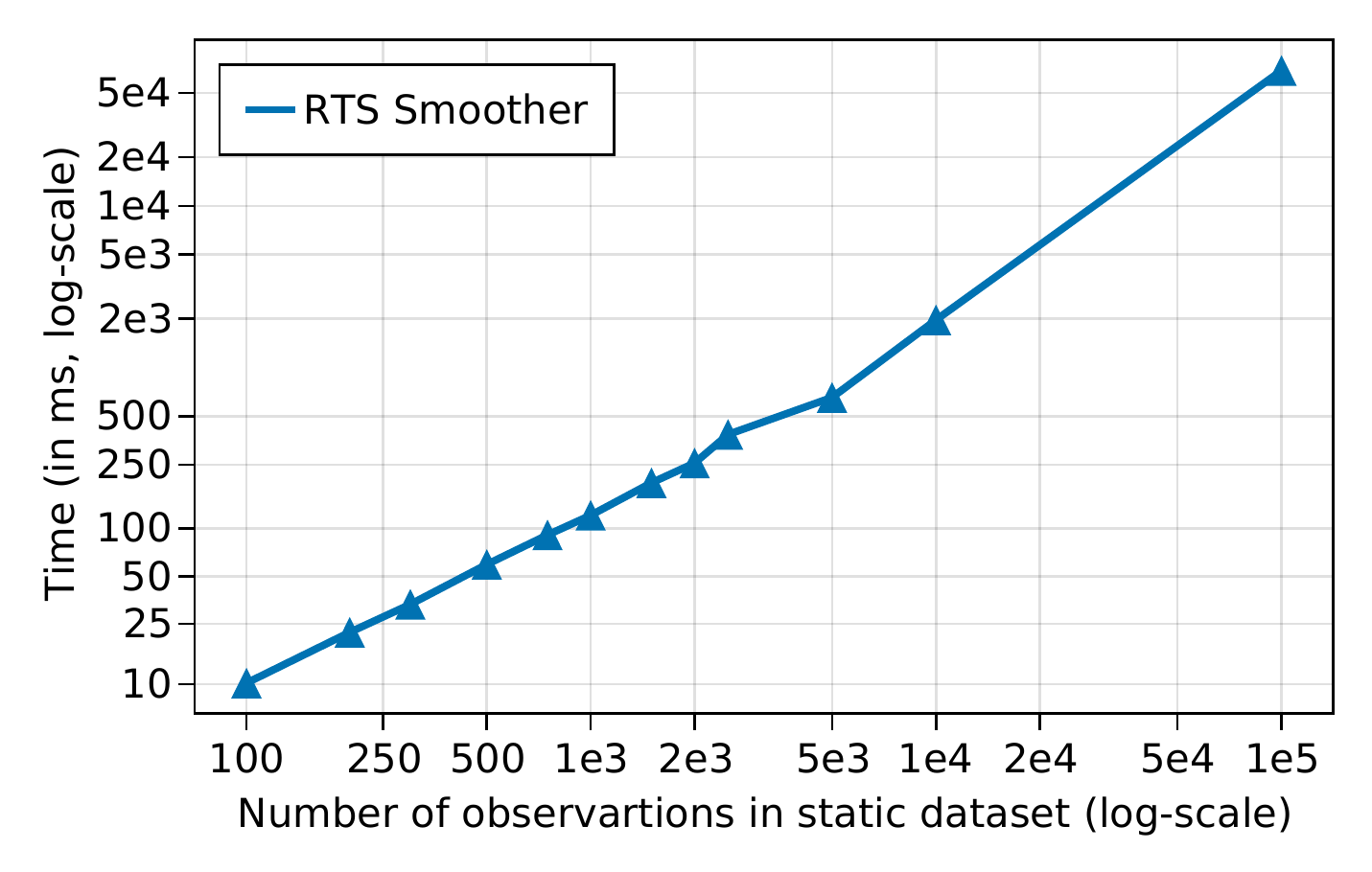}
    \caption{Benchmark scalability results. The x-axis represents the number of observations in dataset and the y-axis corresponds to the absolute execution timings in the benchmark.}
    \label{fig:linear_gaussian_ssm_benchmark}
  \end{subfigure}
  \hspace*{\fill}%
  \caption{Reactive message passing-based inference and benchmark results for the linear Gaussian state space model \eqref{eq:linear_gaussian_ssm} with multivariate (2-dimensional) observations.}
  \label{fig:linear_gaussian_ssm_50}
\end{figure}

Bayesian inference in this type of model can be performed through filtering and smoothing. Although many different variants exist, in our example, we focus on the Rauch-Tung-Striebel (RTS) smoother \citep{sarkka_unscented_2008}, which can be interpreted as performing Belief Propagation algorithm on the full model graph. We show the model specification of this example in Listing~\ref{code:lgssm_model_specification_smoothing}. Fig.~\ref{fig:linear_gaussian_ssm_50} shows the recovered hidden states using the ReactiveMP.jl package. Code examples for the Kalman filter inference procedure for this type of model can be found in the RMP experiments repository on GitHub.

\begin{jllisting}[label={code:lgssm_model_specification_smoothing}, language=julia, style=jlcodestyle, caption={An example of model specification for the linear Gaussian state space model \eqref{eq:linear_gaussian_ssm}.},captionpos=b,float,floatplacement=H]
@model function linear_gaussian_state_space_model(n, d, A, B, P, Q)
  # We create a sequence of random variables of length `n`
  x = randomvar(n) 
  
  # We create a sequence of observed variables of length `n`
  y = datavar(Vector{Float64}, n) 
    
  # Prior distribution for x[1], `d` is the dimension of observations
  # Here we use mean-covariance parametrisation 
  # for Gaussian distribution, but it is also possible to use different 
  # parametrisations such as mean-precision or weighted-mean-precision
  x[1] ~ MvGaussianMeanCovariance(zeros(d), 100.0 * diageye(d)) 
  y[1] ~ MvGaussianMeanCovariance(B * x[1], Q)
  
  for t in 2:n
      x[t] ~ MvGaussianMeanCovariance(A * x[t - 1], P)
      y[t] ~ MvGaussianMeanCovariance(B * x[t], Q)    
  end
  
  # We return `x` and `y` for later reference 
  return x, y
end
\end{jllisting}

\paragraph{Benchmark}

\begin{table}
  \centering
  \begin{tabular}{ |l|r|r|r|r|r|  }
    \hline
    & \multicolumn{5}{|c|}{Number of observations (2-dimensional)} \\
    \hline
    Package & 100 & 200 & 300 & 10\,000 & 100\,000 \\
    \hline
    ReactiveMP.jl & 10 & 22 & 33 & 1971 & 68765\\
    \hline
    ForneyLab.jl (compilation) & 112753 & 374830 & 1057691 & - & -\\ 
    ForneyLab.jl (inference) & 4 & 6 & 12 & - & -\\              
    \hline
    Turing.jl HMC (1000) & 3389 & 7781 & 15822 & - & -\\
    Turing.jl HMC (2000) & 7302 & 15882 & 35545 & - & -\\
    \hline
  \end{tabular}
  \caption{Comparison of run-time duration in milliseconds for automated Bayesian inference for linear Gaussian state space model \eqref{eq:linear_gaussian_ssm} across different packages. Values in the table show minimum possible duration across multiple runs. In this benchmark, the dimension of observations is set to 2. ReactiveMP.jl and ForneyLab.jl use the message passing-based RTS smoothing algorithm on full graph. The ReactiveMP.jl timings include graph creation time. The ForneyLab.jl pipeline consists of model compilation, followed by actual inference execution. Turing.jl uses HMC sampling with 1000 and 2000 number of samples respectively.}
  \label{table:ssm_performance_comparison_2}
\end{table}

\begin{table}
  \centering
  \begin{tabular}{ |l|r|r|r|r|r|  }
    \hline
    & \multicolumn{5}{|c|}{Number of observations (4-dimensional)} \\
    \hline
    Package & 100 & 200 & 300 & 10\,000 & 100\,000 \\
    \hline
    ReactiveMP.jl & 11 & 26 & 40 & 2152 & 85243\\
    \hline
    ForneyLab.jl (compilation) & 112689 & 376606 & 1004868 & - & -\\ 
    ForneyLab.jl (inference) & 4 & 8 & 13 & - & -\\
    \hline
    Turing.jl HMC (1000) & 11246 & 28969 & 61575 & - & -\\
    Turing.jl HMC (2000) & 21337 & 57774 & 116967 & - & -\\
    \hline
  \end{tabular}
  \caption{Comparison of run-time duration in milliseconds for automated Bayesian inference for linear Gaussian state space model \eqref{eq:linear_gaussian_ssm} across different packages. Values in the table show minimum possible duration across multiple runs. In this benchmark dimension of observations is set to 4. ReactiveMP.jl and ForneyLab.jl use message passing-based RTS smoothing algorithm on full graph. The ReactiveMP.jl timings include graph creation time. The ForneyLab.jl pipeline consists of model compilation, followed by actual inference execution. Turing.jl uses HMC sampling with 1000 and 2000 number of samples respectively.}
  \label{table:ssm_performance_comparison_4}
\end{table}

\begin{table}
  \centering
  \begin{tabular}{ |l|r|r|r|r|r|r|  }
    \hline
    & \multicolumn{6}{|c|}{Number of observations}\\
    \hline
    & \multicolumn{3}{|c|}{2-dimensional} & \multicolumn{3}{|c|}{4-dimensional} \\
    \hline
    Method & 100 & 200 & 300 & 100 & 200 & 300\\
    \hline
    Message passing & 3.45 & 3.38 & 3.30 & 6.75 & 6.62 & 6.58\\
    \hline
    HMC (1000) & 6.21 & 11.33 & 26.49 & 15.17 & 24.07 & 42.53\\
    HMC (2000) & 4.62 & 6.24 & 10.54 & 10.02 & 12.76 & 18.27\\
    \hline
  \end{tabular}
  \caption{Comparison of posterior results accuracy in terms of metric \eqref{eq:average_mse} in the linear Gaussian state space model \eqref{eq:linear_gaussian_ssm} across different packages. Lower values indicate better performance. ReactiveMP.jl and ForneyLab.jl use message passing-based RTS smoothing algorithm on full graph. Turing.jl uses HMC sampling with 1000 and 2000 number of samples respectively.}
  \label{table:ssm_accuracy_comparison_2_4}
\end{table}

The main benchmark results are presented in Tables \ref{table:ssm_performance_comparison_2} and \ref{table:ssm_performance_comparison_4}. Accuracy results in terms of the average error metric of \eqref{eq:average_mse} are presented in Table \ref{table:ssm_accuracy_comparison_2_4}. The new reactive message passing-based implementation for Bayesian inference shows better performance and scalability results and outperforms the compared packages in terms of time and memory consumption (not present in the table) significantly. Furthermore, the ReactiveMP.jl package is capable of running inference in very large models with hundreds of thousands of variables. Accuracy results in Table~\ref{table:ssm_accuracy_comparison_2_4} show that message passing-based methods give more accurate results in comparison to sampling-based methods, which is expected as message passing performs exact Bayesian inference in this type of model.

Table~\ref{table:ssm_performance_comparison_2} shows that the ForneyLab.jl package actually executes the inference task for this model faster than the ReactiveMP.jl package. The ForneyLab.jl package analyses the TFFG thoroughly during precompilation and is able to create an efficient predefined message update schedule ahead of time that is able to execute the inference procedure very fast. Unfortunately, as we discussed in the Section \ref{section:motivation}, ForneyLab.jl's schedule-based solution suffers from long latencies in the graph creation and precompilation stages. This behavior limits investigations of "what-if"-scenarios in the model specification space. On the other hand, the ReactiveMP.jl creates and executes the reactive message passing scheme dynamically without full graph analysis. This strategy helps with scalability for very large models but comes with some run-time performance costs.


The execution timings of the Turing.jl package and the corresponding HMC algorithm mostly depend on the number of samples in the sampling procedure and other hyper parameters, which usually need to be fine tuned. In general, in sampling-based packages, a larger number of samples leads to better approximations but longer run times. On the other hand, as we mentioned in the beginning of the Section \ref{section:experiments}, the Turing.jl platform and its built-in algorithms support running inference in a broader class of probabilistic models, as it does not depend on analytical solutions and is not restricted to work with closed-form update rules. 

We present benchmark results for $10\,000$ and $100\,000$ observations only for the ReactiveMP.jl package since for other compared packages it involves running the inference for more than an hour. For example, we can estimate that for a static dataset with $100\,000$ observations the corresponding TFFG for this model has roughly $400\,000$ nodes. The ReactiveMP.jl executes inference for such a large model under two minutes. Therefore, we conclude that the new reactive message passing implementation scales better with the number of random variables and is able to run efficient Bayesian inference for large conjugate state space models.

\subsection{Hidden Markov Model}\label{section:experiment_hmm}

In this example, the goal is to perform Bayesian inference in a Hidden Markov model (HMM). An HMM can be viewed as a specific instance of a state space model in which the latent variables are discretely valued. HMMs are widely used in speech recognition \citep{jelinek_statistical_1998, rabiner_fundamentals_1993}, natural language modelling \citep{manning_foundations_1999} and in many other related fields.

We consider an HMM specified by
\begin{subequations}\label{eq:hmm_equations}
  \begin{align}
    p(A) &= \text{MatrixDirichlet}(A|P_A),\\
    p(B) &= \text{MatrixDirichlet}(B|P_B),\\
    p(z_t|z_{t-1}) &= \text{Cat}(z_t|Az_{t - 1})\label{eq:hmm_equations-state},\\
    p(y_t|z_t) &= \text{Cat}(y_t|Bz_{t}),
  \end{align}
\end{subequations} 
where $A$ and $B$ are state transition and observation model matrices respectively, $z_t$ is a discrete $M$-dimensional one-hot coded latent state, $y_t$ is the observation at time step $t$, $\text{MatrixDirichlet}(\;\cdot\;|\;P)$ denotes a matrix variate generalisation of $\text{Dirichlet}$ distribution with concentration parameters matrix $P$ \citep{zhang_matrix-variate_2014} and $\text{Cat}(\;\cdot\;|\;p)$ denotes a Categorical distribution with concentration parameters vector $p$. One-hot coding of $z_t$ implies that $z_{t,i} \in \{0,1\}$ and $\sum_{i=1}^M z_{t,i} = 1$. With this encoding scheme, \eqref{eq:hmm_equations-state} is short-hand for
\begin{equation}
 p(z_{t,i} = 1|z_{t-1,j} = 1) = A_{ij} \,.   
\end{equation}

\paragraph{CBFE Specification}

Exact Bayesian inference for this model is intractable and we resort to approximate inference by message passing-based minimization of CBFE. For the variational family of distributions $\mathcal{Q}_B$ we assume a structured factorisation around state transition probabilities and the mean-field factorisation assumption for every other factor in the model:

\begin{subequations}\label{eq:hmm_recognition_factorisation}
\begin{align}
     q(z, A, B) &= q(z) q(A) q(B), \\
     q(z) &= \frac{\prod_{t = 2}^{T} q(z_{t - 1}, z_t)}{\prod_{t = 2}^{T} q(z_t) }.
\end{align} 
\end{subequations}

\begin{figure}[h]
  \centering
  \hspace*{\fill}%
  \begin{subfigure}[t]{.45\textwidth}
    \includegraphics[width=\columnwidth]{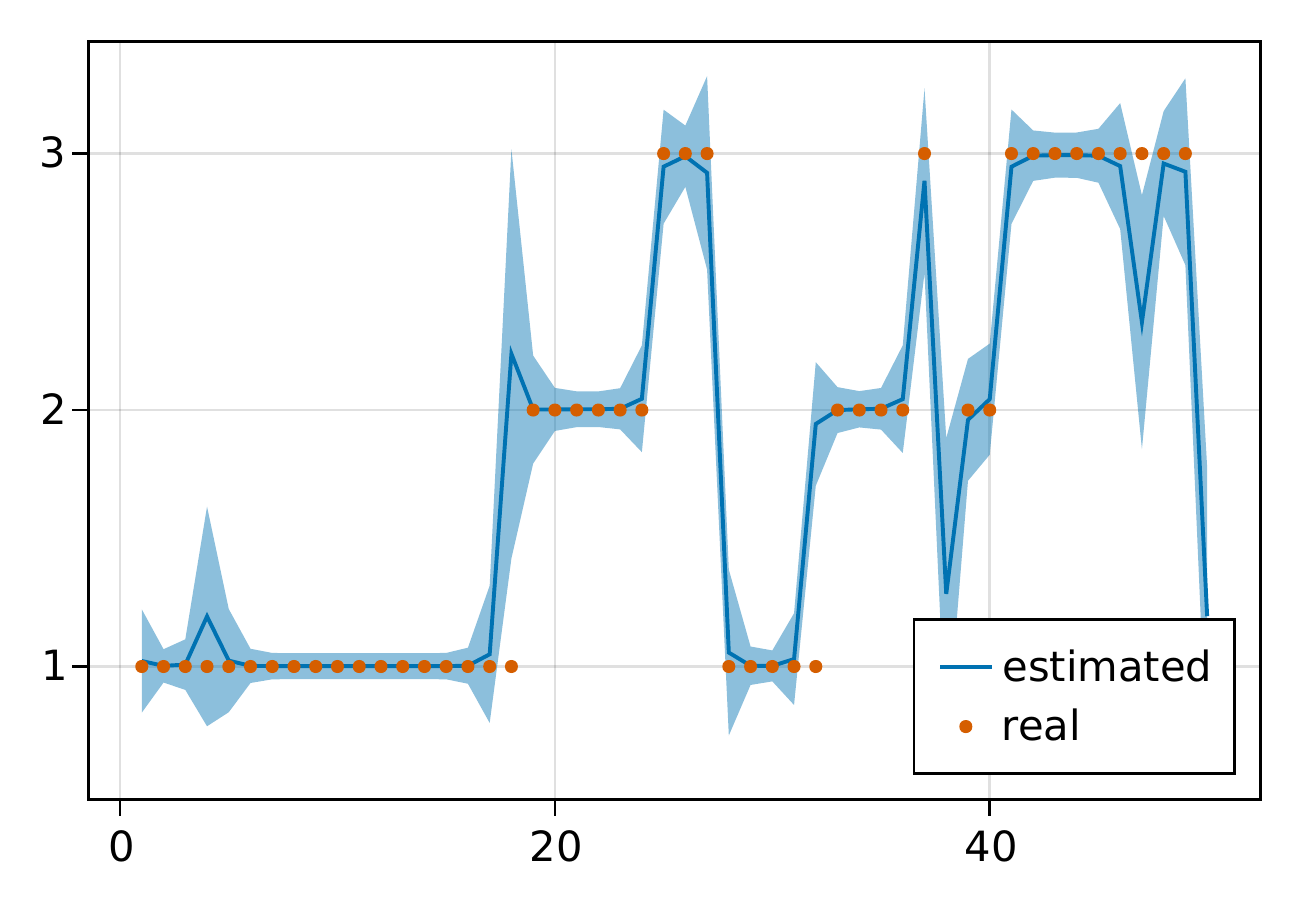}
    \caption{Hidden states inference results. The orange dots represent real values of states at each time step. The blue line represents the mean of the posterior over latent states with one standard deviation. }
    \label{fig:hmm_inference}
  \end{subfigure}
  \hspace*{\fill}%
  \begin{subfigure}[t]{.45\textwidth}
    \includegraphics[width=\columnwidth]{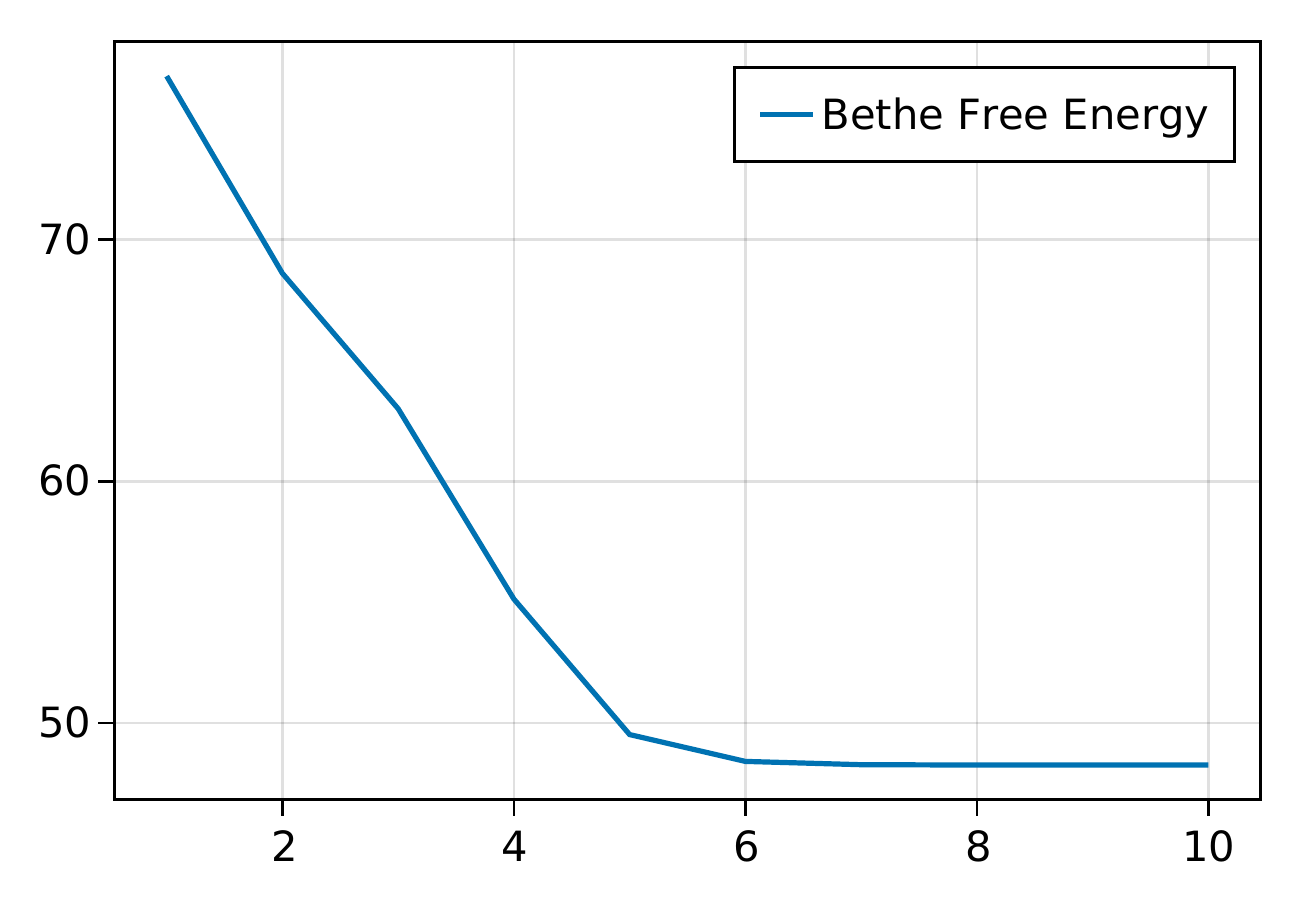}
    \caption{Bethe Free Energy evaluation results. The x-axis represents an index of VMP iteration. The y-axis represents a Bethe Free Energy value at a specific VMP iteration.}
    \label{fig:hmm_fe}
  \end{subfigure}
  \hspace*{\fill}%
  \caption{Inference results for the Hidden Markov model in Listing \ref{code:hmm_model_specification}.}
  \label{fig:hmm_inference_results}
\end{figure}

We show the model specification code for this model with the extra factorisation constraints \eqref{eq:hmm_recognition_factorisation} specification in Listing~\ref{code:hmm_model_specification} and inference results in Fig.~\ref{fig:hmm_inference_results}. Fig.~\ref{fig:hmm_fe} shows convergence of the BFE values after several number of VMP iterations. Qualitative results in Fig.~\ref{fig:hmm_inference} show reasonably correct posterior estimation of discretely valued hidden states.

\begin{jllisting}[label={code:hmm_model_specification}, language=julia, style=jlcodestyle, caption={An example of model specification for the Hidden Markov model \eqref{eq:hmm_equations}.},captionpos=b,float,floatplacement=H]
# `@model` macro accepts an optional list of default parameters
# Here we use `MeanField` factorisation assumption by default for every node
@model [ default_factorisation = MeanField() ]
function hidden_markov_model(n, priorA, priorB)
  
  A ~ MatrixDirichlet(priorA)
  B ~ MatrixDirichlet(priorB)
    
  z = randomvar(n)
  y = datavar(Vector{Float64}, n)
  
  # `Transition` node is an alias for `Categorical(B * z[ ])`
  z[1] ~ Categorical(fill(1.0 / 3.0, 3))
  y[1] ~ Transition(z[1], B)
    
  for t in 2:n
    # We override the default `MeanField` assumption with 
    # structured posterior factorisation assumption using `where` block
    z[t] ~ Transition(z[t - 1], A) where { q = q(z[t - 1], z[t])q(A) }
    y[t] ~ Transition(z[t], B)
  end
  
  return A, B, z, y
end
\end{jllisting}

\paragraph{Benchmark}

\begin{figure}[h]
  \centering
  \hspace*{\fill}%
  \begin{subfigure}[t]{.45\textwidth}
    \includegraphics[width=\columnwidth]{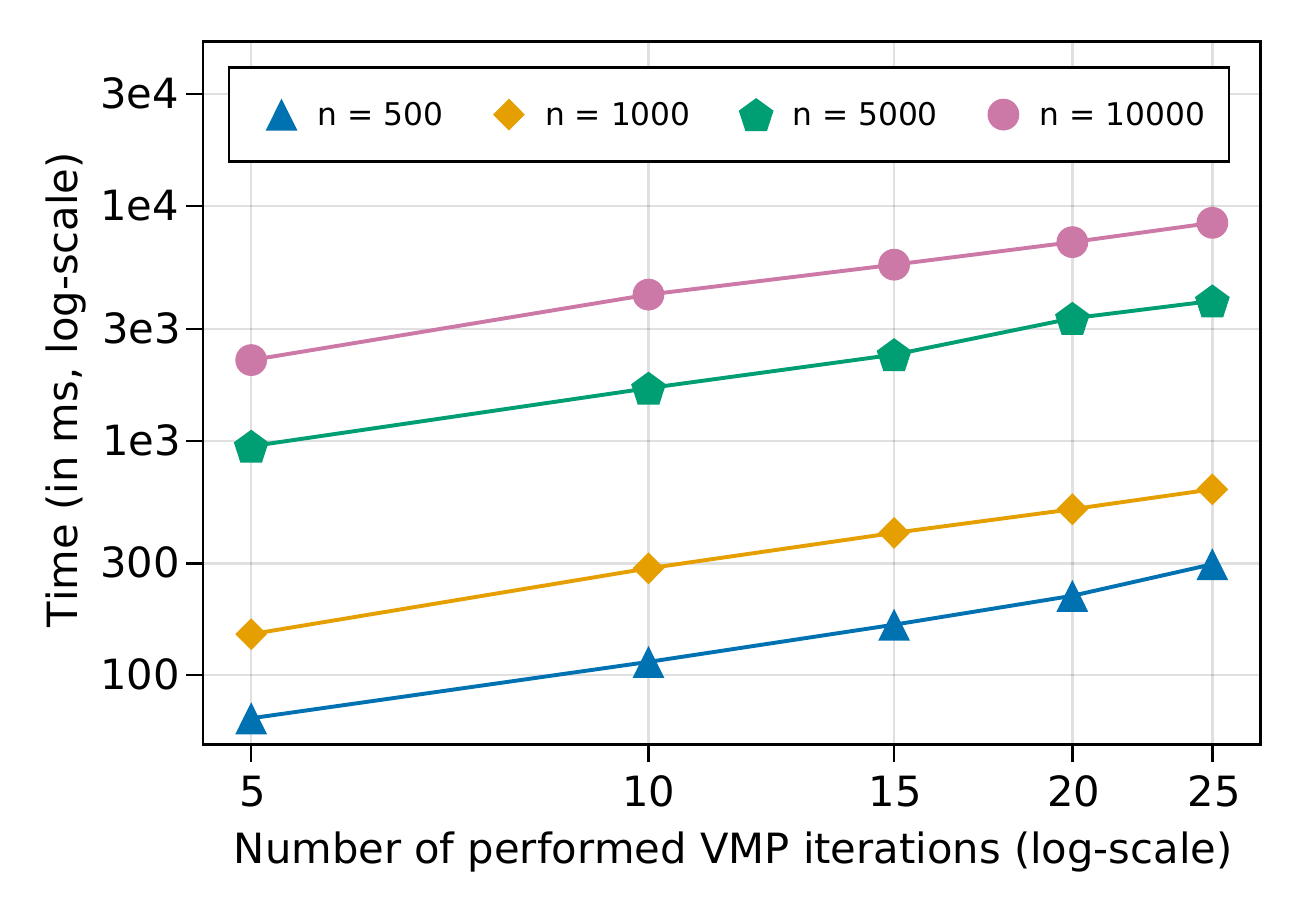}
    \caption{Run-time duration vs number of VMP iterations for different number of observations in the data set $n$.}
    \label{fig:hmm_benchmark_iterations}
  \end{subfigure}
  \hspace*{\fill}%
  \begin{subfigure}[t]{.45\textwidth}
    \includegraphics[width=\columnwidth]{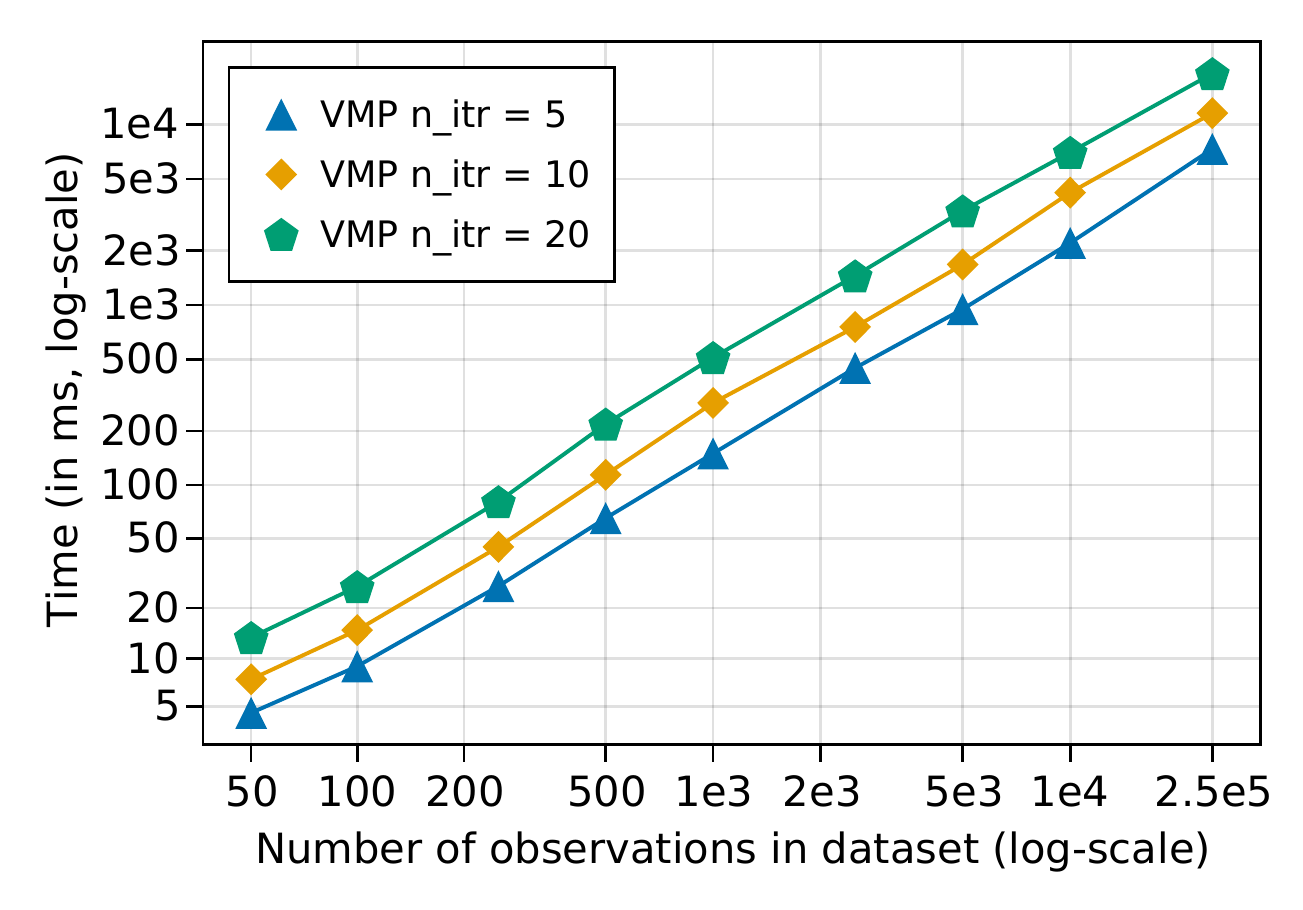}
    \caption{Run-time duration vs number of observations in the data set for a different number of VMP iterations $\mathrm{n\_itr}$.}
    \label{fig:hmm_benchmark_observations}
  \end{subfigure}
  \hspace*{\fill}%
  \caption{Benchmark results for the Hidden Markov model in Listing \ref{code:hmm_model_specification}.}
  \label{fig:hmm_benchmark}
\end{figure}

The main benchmark results are presented in Table~\ref{table:hmm_performance_comparison}. Fig.~\ref{fig:hmm_benchmark} presents a comparison of the performance of the ReactiveMP.jl as a function of the number of VMP iterations and shows that the resulting inference procedure scales linearly both on the number of observations and the number of VMP iterations performed. In the context of VMP, each iteration decreases the Bethe Free and effectively leads to a better approximation for the marginals over the latent variables. As in our previous example, we show the accuracy results for message passing-based methods in comparison with sampling-based methods in terms of metric \eqref{eq:average_mse} in Table~\ref{table:hmm_accuracy_comparison}. The ForneyLab.jl shows the same level of posterior accuracy, as it uses the same VMP algorithm, but is slower in model compilation and execution times. The Turing.jl is set to use the HMC method for estimating the posterior of transition matrices $A$ and $B$ with a combination of particle Gibbs (PG) for discrete states $z$.

\begin{table}
  \centering
  \begin{tabular}{ |l|r|r|r|r|  }
    \hline
    & \multicolumn{4}{|c|}{Number of observations} \\
    \hline
    Package & 100 & 250 & 10\,000 & 25\,000 \\
    \hline
    ReactiveMP.jl & 21 & 62 & 5624 & 15544\\
    \hline
    ForneyLab.jl (compilation) & 82280 & 405949 & - & -\\
    ForneyLab.jl (inference) & 93 & 228 & - & -\\
    \hline
    Turing.jl HMC+PG (250) & 93956 & 396946 & - & -\\
    Turing.jl HMC+PG (500) & 186293 & 783850 & - & -\\
    \hline
  \end{tabular}
  \caption{Comparison of run-time duration in milliseconds for automated Bayesian inference for Hidden Markov model \eqref{eq:hmm_equations} across different packages. Values in the table show minimum possible duration across multiple runs. In this benchmark the number of categories $M$ of observations is set to 3. ReactiveMP.jl and ForneyLab.jl perform VMP on a full graph. The number of VMP iterations is set to 15. The ReactiveMP.jl timings include graph creation time.  The ForneyLab.jl pipeline consists of model compilation, followed by actual inference execution. Turing.jl uses a combination of HMC sampling and particle Gibbs sampling and runs two benchmarks with 250 and 500 number of samples respectively.}
  \label{table:hmm_performance_comparison}
\end{table}

\begin{table}
  \centering
  \begin{tabular}{ |l|r|r|r| }
    \hline
    & \multicolumn{3}{|c|}{Number of observations} \\
    \hline
    Method & 50 & 100 & 250 \\
    \hline
    Message passing & 0.10 & 0.09 & 0.07 \\
    \hline
    HMC+PG (250) & 0.50 & 0.51 & 0.52\\
    HMC+PG (500) & 0.51 & 0.49 & 0.51\\
    \hline
  \end{tabular}
  \caption{Comparison of posterior results accuracy in terms of metric \eqref{eq:average_mse} in the Hidden Markov model \eqref{eq:hmm_equations} across different packages. Lower values indicate better performance. In this experiment the number of categories $M$ of observations is set to 3. ReactiveMP.jl and ForneyLab.jl perform VMP on a full graph. The number of VMP iterations is set to 15. Turing.jl uses a combination of HMC sampling and particle Gibbs sampling and runs two benchmarks with 250 and 500 number of samples respectively.}
  \label{table:hmm_accuracy_comparison}
\end{table}

\subsection{Hierarchical Gaussian Filter}\label{section:experiment_hgf}

For our last example, we consider Bayesian inference in the Hierarchical Gaussian Filter (HGF) model. The HGF is popular in the computational neuroscience literature and is often used for the Bayesian modelling of volatile environments, such as uncertainty in perception or financial data \citep{mathys_uncertainty_2014}. The HGF is essentially a Gaussian random walk, where the time-varying variance of the random walk is determined by the state of a higher level process, e.g., a Gaussian random walk with fixed volatily or another more complex signal. Specifically, a simple HGF is defined as

\begin{subequations}\label{eq:hgf}
    \begin{align}
    p(s^{(j)}_t|s^{(j)}_{t-1}) &= \,\mathcal{N}(s^{(j)}_k|s^{(j)}_{t - 1}, f(s^{(j + 1)}_t))\,, \text{for }j=1,2,\ldots,J\label{eq:hgf_nonlinear_mapping} \\
    p(y_t|s^{(1)}_t) &= \,\mathcal{N}(y_t|s^{(1)}_t, \tau)
  \end{align}
\end{subequations} 
where $y_t$ is the observation at time step $t$, $s_t^{(j)}$ is the latent state for the $j$-th layer at time step $t$, and $f$ is a link function, which maps states from the (next higher) $(j+1)-th$ layer to the non-negative variance of the random walk in the $j$-th layer. The HGF model typically defines the link function as $f(s^{(j + 1)}_t) = \exp(\kappa s^{(j + 1)}_t + \omega)$, where $\kappa$ and $\omega$ are either hyper-parameters or random variables included in the model.

As in the previous example, the exact Bayesian inference for this type of model is not tractable. Moreover, the variational message update rules \eqref{eq:node_outbound_message_update_vmp} are not tractable either since the HGF model contains non-conjugate relationships between variables in the form of the link function $f$. However, inference in this model is still possible with custom message update rules approximation \citep{senoz_online_2018}. The ReactiveMP.jl package supports a straightforward API to add custom nodes with custom message passing update rules. In fact, a big part of the ReactiveMP.jl library is a collection of well-known factor nodes and message update rules implemented using the same API as for custom novel nodes and message update equations. The API also supports message update rules that involve extra nontrivial approximation steps. In this example, we added a custom \textit{Gaussian controlled variance} (GCV) node to model the nonlinear time-varying variance mapping from different hierarchy levels \eqref{eq:hgf_nonlinear_mapping}, with a set of approximate update rules based on the Gauss-Hermite cubature rule from \citep{kokkala_sigma-point_2015}. The main purpose of this example is to show that ReactiveMP.jl is capable of running inference in complex non-conjugate models, but requires creating custom factor nodes and choosing appropriate integral calculation approximation methods. 

\paragraph{CBFE Specification}

In this example, we want to show an example of reactive online Bayesian learning (filtering) in a 2-layer HGF model. For simplicity and to avoid extra clutter, we assume $\tau_k$, $\kappa$ and $\omega$ to be fixed, but there are no principled limitations to make them random variables, endow them with priors and to estimate their corresponding posterior distributions. For online learning in the HGF model \eqref{eq:hgf} we define only a single time step and specify the structured factorisation assumption for state transition nodes as well as for higher layer random walk transition nodes and the mean-field assumption for other functional dependencies in the model:

\begin{subequations}\label{eq:hgf_recognition_factorisation}
\begin{align}
     q(s^{(1)}, s^{(2)}) &= q(s^{(1)}) q(s^{(2)}), \\
     q(s^{(1)}) &= \frac{\prod_{t = 2}^{T} q(s^{(1)}_{t - 1}, s^{(1)}_t)}{\prod_{t = 2}^{T} q(s^{(1)}_t) },\\
     q(s^{(2)}) &= \frac{\prod_{t = 2}^{T} q(s^{(2)}_{t - 1}, s^{(2)}_t)}{\prod_{t = 2}^{T} q(s^{(2)}_t) }.
\end{align} 
\end{subequations}
 
We show an example of a HGF model specification in Listing \ref{code:hgf_model}. For the approximation of the integral in the message update rule, the \jlinl{GCV} node requires a suitable approximation method to be specified during model creation. For this reason we create a \jlinl{metadata} object called \jlinl{GCVMetadata()} that accepts a \jlinl{GaussHermiteCubature()} approximation method with a prespecified number of sigma points \jlinl{gh_n}.

\begin{jllisting}[language=julia, style=jlcodestyle, label={code:hgf_model}, caption={An example of model specification for the Hierarchical Gaussian Filter model \eqref{eq:hgf}.},captionpos=b,float,floatplacement=H]
# We use `MeanField` factorisation assumption by default for every node
@model [ default_factorisation = MeanField() ] 
function hierarchical_gaussian_filter_model(gh_n, s_2_w, y_w, kappa, omega)
  # `s_2` refers to the second layer in the hierarchy
  # `s_1` refers to the first layer in the hierarchy
  s_2_prior = datavar(Float64, 2)
  s_1_prior = datavar(Float64, 2)
  y         = datavar(Float64)
  
  s_2_previous ~ GaussianMeanPrecision(z_2_prior[1], z_2_prior[2])
  s_1_previous ~ GaussianMeanPrecision(s_1_prior[1], s_1_prior[2])
  
  # Z-layer modelled as a random walk with structured factorisation assumption
  s_2 ~ GaussianMeanPrecision(s_2_previous, s_2_w) where {
    q = q(s_2_previous, s_2)q(s_2_w)
  }
  
  # GCV node uses Gauss-Hermite cubature to approximate the nonlinearity 
  # between layers in the hierarchy. We may change the number of points 
  # used in the approximation with the `gh_n` model argument
  meta = GCVMetadata(GaussHermiteCubature(gh_n))
  
  # Comma syntax for the tilde operator allows us to extract a reference to 
  # the GCV node, which we will use later on to initialise joint marginals
  gcv, s_1 ~ GCV(s_1_previous, s_2, kappa, omega) where { 
    q = q(s_1, s_1_previous)q(s_2)q(kappa)q(omega), meta = meta 
  }
  
  y ~ GaussianMeanPrecision(s_1, y_w)
  
  return s_2_prior, s_1_prior, gcv, s_2, s_1, y
end
\end{jllisting}

\paragraph{Inference}

For the inference procedure, we adopt the technique from the Infinite Recursive Chain Processing section (Section \ref{section:infinite_reaction_chain_processing}). We subscribe to future observations, perform a prespecified number of VMP iterations for each new data point, and redirect the last posterior update as a prior for the next future observation. The resulting inference procedure reacts to new observations autonomously and is compatible with infinite data streams. We show a part of the inference procedure for the model \eqref{eq:hgf} in Listing~\ref{code:hgf_inference}. Full code is available at GitHub experiment's repository.

\begin{jllisting}[language=julia, style=jlcodestyle, label={code:hgf_inference}, caption={A part of an example of inference specification for the Hierarchical Gaussian Filter model \eqref{eq:hgf}.},captionpos=b,float,floatplacement=H]
# This function will be called every time we observe a new data point
function on_next!(actor::HGFInferenceActor, data::Float64)
  s_2_prior, s_1_prior, gcv, s_2, s_1, y = actor.model_output
  
  # To perform multiple VMP iterations we pass the data multiple times
  # It forces the inference backend to react to the data and to
  # update posterior marginals multiple times
  for i in 1:actor.n_vmp_iterations
      update!(s_2_prior, actor.current_s_2)
      update!(s_1_prior, actor.current_s_1)
      update!(y, data)
  end
  ...
  # Update current posterior marginals at time step `t`
  actor.current_s_2 = mean_precision(last(actor.history_s_2))
  actor.current_s_1 = mean_precision(last(actor.history_s_1))
end
\end{jllisting}

\begin{figure}[h]
  \centering
  \hspace*{\fill}%
  \begin{subfigure}[t]{.31\textwidth}
    \includegraphics[width=\columnwidth]{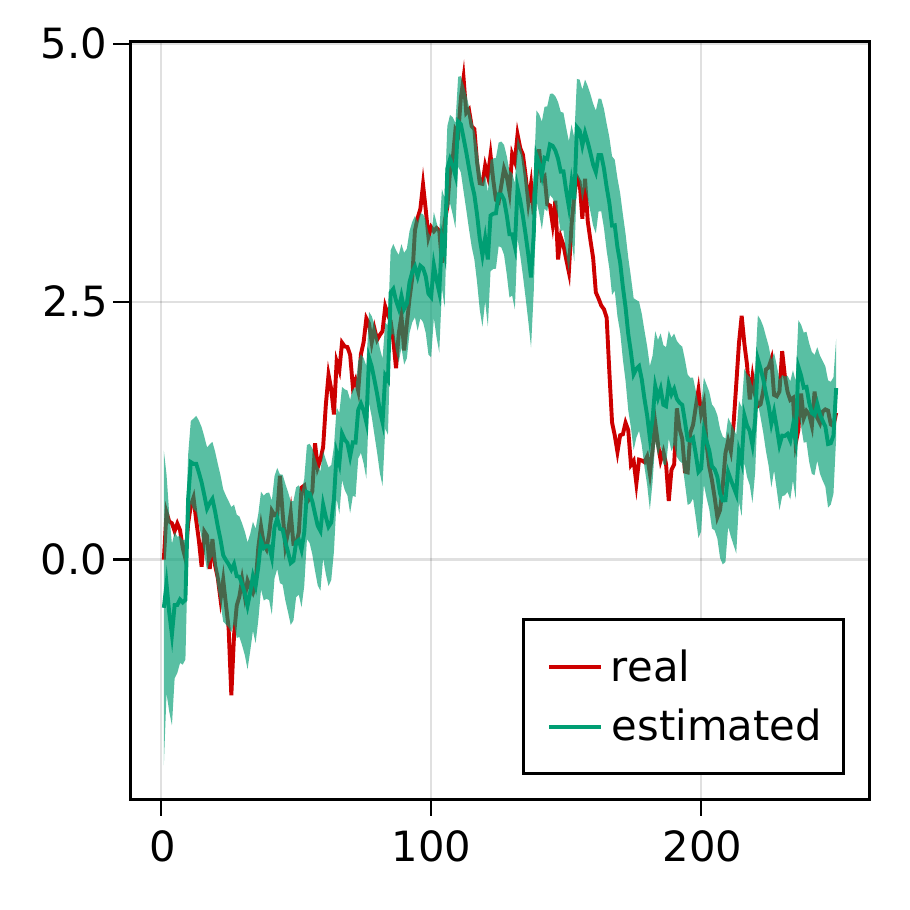}
    \caption{The inference results of layer $s_t^{(2)}$ hidden states. The x-axis represents the time steps and the y-axis corresponds to the actual value of hidden states.}
    \label{fig:hgf_layer_z}
  \end{subfigure}
  \hspace*{\fill}%
  \begin{subfigure}[t]{.31\textwidth}
    \includegraphics[width=\columnwidth]{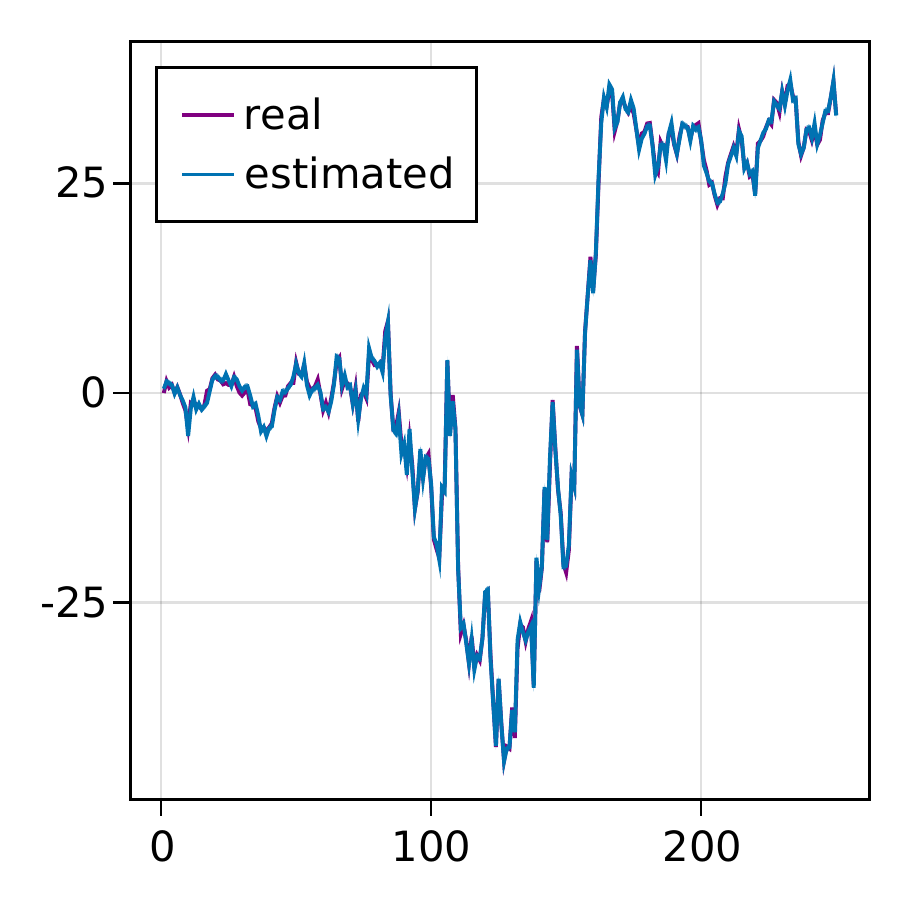}
    \caption{The inference results of layer $s_t^{(1)}$ hidden states. The x-axis represents the time steps and the y-axis corresponds to the actual value of hidden states.}
    \label{fig:hgf_layer_x}
  \end{subfigure}
  \hspace*{\fill}%
  \begin{subfigure}[t]{.31\textwidth}
    \includegraphics[width=\columnwidth]{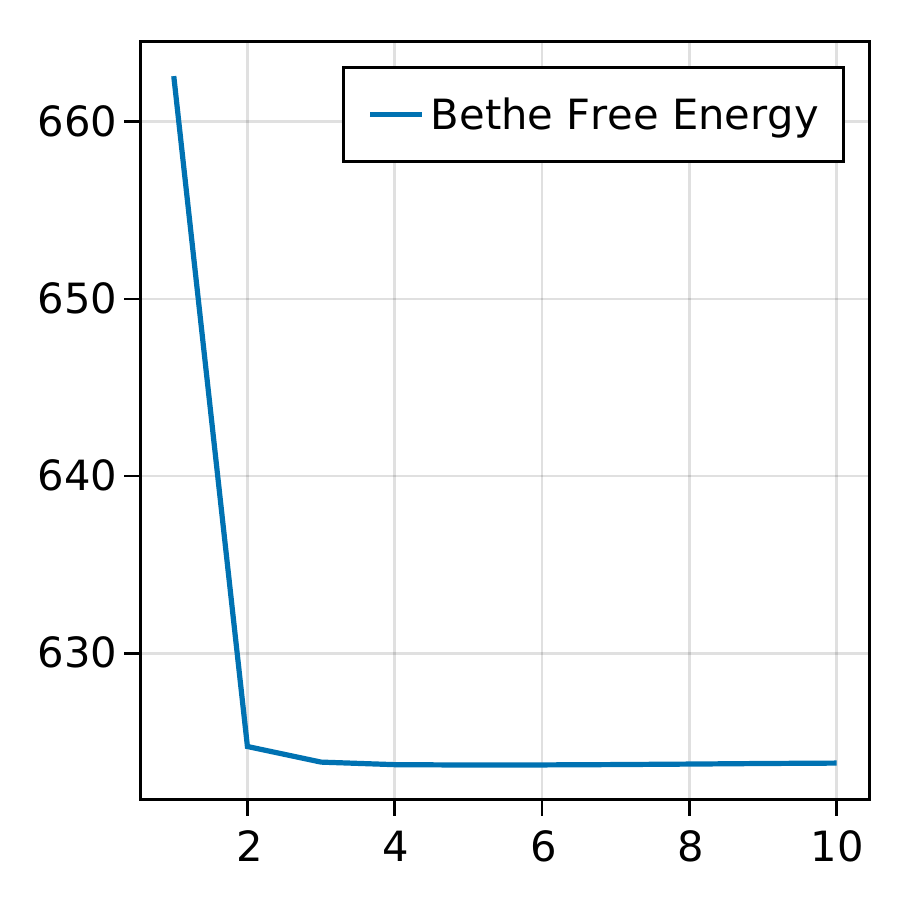}
    \caption{Bethe Free Energy evaluation results. The x-axis represents an index of VMP iteration. The y-axis represents a Bethe Free Energy value averaged over all data points at a specific VMP iteration.}
    \label{fig:hgf_fe}
  \end{subfigure}
  \hspace*{\fill}%
  \caption{
    Online learning inference results for the Hierarchical Gaussian Filter model \eqref{eq:hgf} for 250 synthetically generated 1-dimensional observations. 
    }
  \label{fig:hgf_filtering_inference}
\end{figure} 

We show the inference results in Fig.~\ref{fig:hgf_filtering_inference}. Fig.~\ref{fig:hgf_fe} shows convergence of the Bethe Free Energy after several number of VMP iterations. Qualitative results in Fig.~\ref{fig:hgf_layer_z} and in Fig.~\ref{fig:hgf_layer_x} show correct posterior estimation of continuously valued hidden states even though the model contains non-conjugate relationships between variables.

\paragraph{Benchmark}

\begin{figure}[h]
  \centering
  \hspace*{\fill}%
  \begin{subfigure}[t]{.43\textwidth}
    \includegraphics[width=\columnwidth]{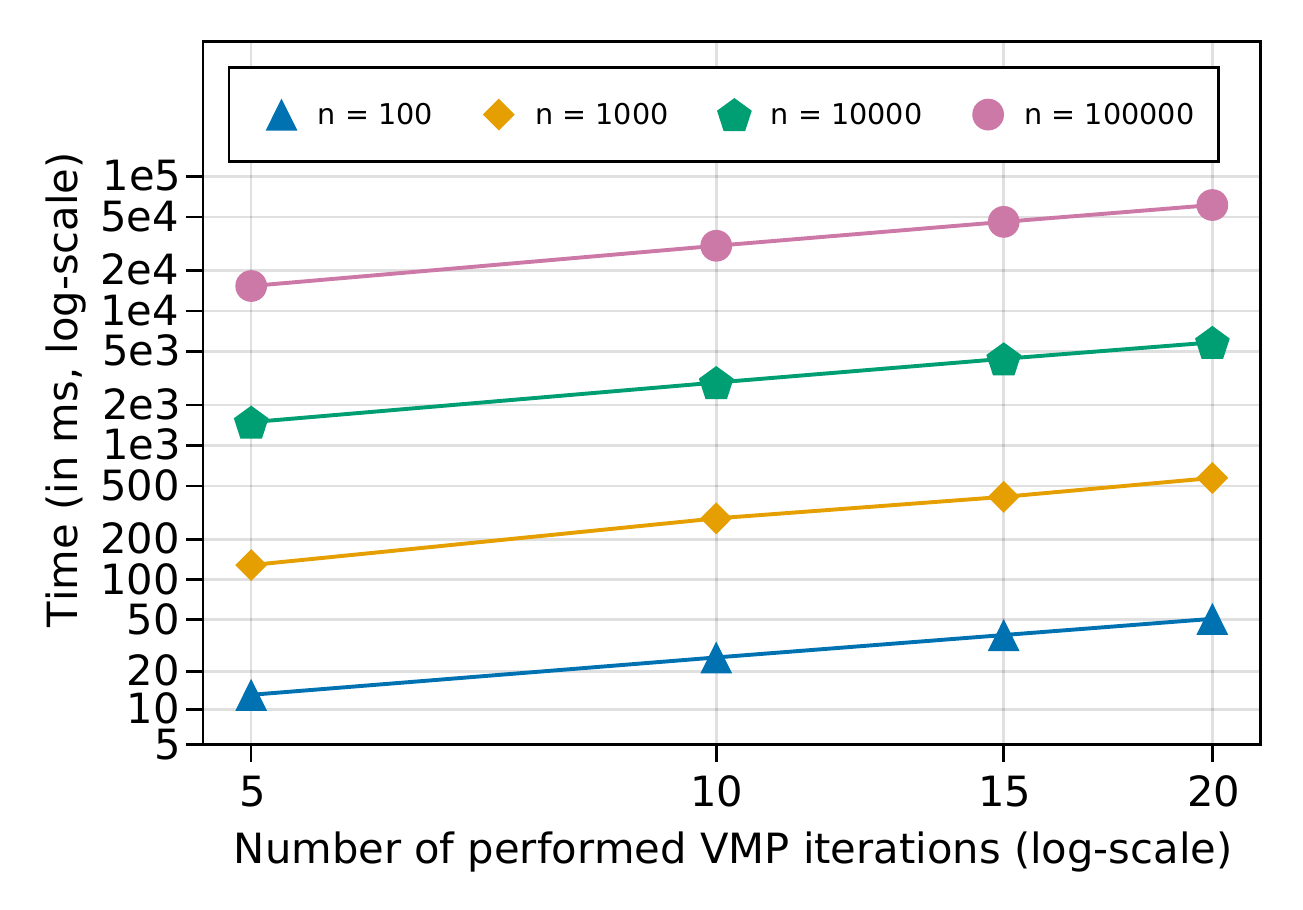}
    \caption{Benchmark: run-time duration vs number of VMP iterations for different number of observations in dataset $n$.}
    \label{fig:hgf_benchmark_iterations}
  \end{subfigure}
  \hspace*{\fill}%
  \begin{subfigure}[t]{.43\textwidth}
    \includegraphics[width=\columnwidth]{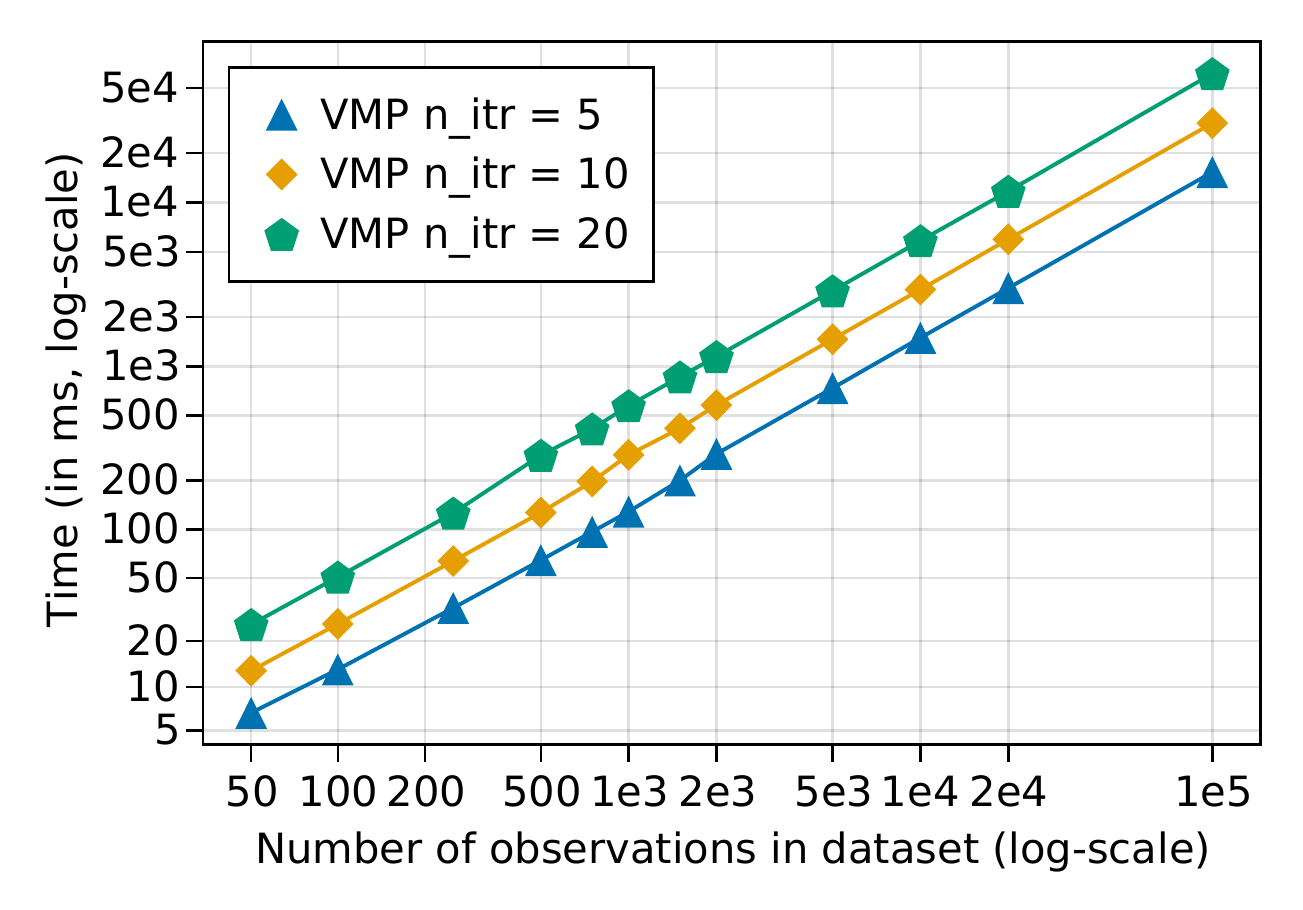}
    \caption{Benchmark: run-time duration vs number of observations in the data set for different number of VMP iterations $\mathrm{n\_itr}$.}
    \label{fig:hgf_benchmark_observations}
  \end{subfigure}
  \hspace*{\fill}%
  \caption{Benchmark results for the Hierarchical Gaussian Filter Model \eqref{eq:hgf} in Listing \ref{code:hgf_model}.}
  \label{fig:hgf_benchmark}
\end{figure}

The main benchmark results are presented in Table~\ref{table:hgf_performance_comparison} and accuracy comparison in Table~\ref{table:hgf_accuracy_comparison}. We show the performance of the ReactiveMP.jl package depending on the number of observations and the number of VMP iterations for this particular model in Fig.~\ref{fig:hgf_benchmark}. As in the previous example, we note that that the ReactiveMP.jl framework scales linearly both with the number of observations and with the number of VMP iterations performed.

\begin{table}
  \centering
  \begin{tabular}{ |l|r|r|r|r|r|  }
    \hline
    & \multicolumn{5}{|c|}{Number of observations} \\
    \hline
    Package & 50 & 100 & 250 & 10\,000 & 100\,000 \\
    \hline
    ReactiveMP.jl & 19 & 38 & 95 & 4412 & 45889\\
    \hline
    ForneyLab.jl (compilation) & 590 & 582 & 592 & - & -\\
    ForneyLab.jl (inference) & 457 & 876 & 2368 & - & -\\
    \hline
    Turing.jl (500) & 1357 & 2607 & 6531 & - & -\\
    Turing.jl (1000) & 3036 & 5287 & 12847 & - & -\\
    \hline
  \end{tabular}
  \caption{Comparison of run-time duration in milliseconds for automated Bayesian inference in the Hierarchical Gaussian Filter model \eqref{eq:hgf} across different packages. Values in the table show minimum possible duration across multiple runs. ReactiveMP.jl and ForneyLab.jl perform online learning with VMP on a single time step of the corresponding graph. The number of VMP iterations performed is set to 15. The ReactiveMP.jl timings include graph creation time. The ForneyLab.jl pipeline consists of model compilation, followed by actual inference execution. Turing.jl uses HMC sampling with 500 and 1000 number of samples respectively.}
  \label{table:hgf_performance_comparison}
\end{table}

\begin{table}
  \centering
  \begin{tabular}{ |l|r|r|r|r|r|r| }
    \hline
    & \multicolumn{6}{|c|}{Number of observations} \\
    \hline
    & \multicolumn{3}{|c|}{$x_k^{(2)}$ layer} & \multicolumn{3}{|c|}{$x_k^{(1)}$ layer} \\
    \hline
    Method & 50 & 100 & 250 & 50 & 100 & 250 \\
    \hline
    Message passing & 0.89 & 0.79 & 0.75 & 0.36 & 0.35 & 0.35 \\
    \hline
    HMC (500) & 1.41 & 1.29 & 1.51 & 0.45 & 0.62 & 4.74\\
    HMC (1000) & 1.30 & 1.14 & 1.22 & 0.41 & 0.49 & 2.56\\
    \hline
  \end{tabular}
  \caption{Comparison of posterior results accuracy in terms of metric \eqref{eq:average_mse} in the Hierarchical Gaussian Filter model \eqref{eq:hgf} across different packages. Lower values indicate better performance. ReactiveMP.jl and ForneyLab.jl perform online learning with VMP on a single time step of the corresponding graph. The number of VMP iterations is set to 15. Turing.jl uses HMC sampling with 500 and 1000 number of samples respectively.}
  \label{table:hgf_accuracy_comparison}
\end{table}

We can see that, in contrast with previous examples where we performed inference on a full graph, the ForneyLab.jl compilation time is not longer dependent on the number of observations and the model compilation is more acceptable due to the fact that we always build a single time step of a graph and reuse it during online learning. Both the ReactiveMP.jl and ForneyLab.jl show the same scalability and posterior accuracy results as they both use the same method for posterior approximation, however, ReactiveMP.jl is faster in VMP inference execution in absolute timing. 

In this model, the HMC algorithm shows less accurate results in terms of the metric \eqref{eq:average_mse}. However, because of the online learning setting, the model has a small number of unknowns, which makes the HMC algorithm more feasible to perform inference for a large number of observations in comparison with the previous examples. The advantage of using Turing.jl is that it supports inference for non-conjugate relationships between variables in a model by default, and, hence, does not require hand-crafted custom factor nodes and message update rule approximations.

\section{Discussion}\label{section:discussion}

We are investigating several possible future directions for the reactive message passing framework that could improve its usability and performance. We believe that the current implementation is efficient and flexible enough to perform real-time reactive Bayesian inference and has been tested on both synthetic and real data sets. The new framework allows for large models to be applied in practice and simplifies the model exploration process in signal processing applications. A natural future direction is to apply the new framework to an \emph{Active Inference} \citep{friston_active_2016} setting. A reactive active inference agent learns purposeful behavior solely through situated interactions with its environment and processing these interactions by real-time inference. As we discussed in the Motivation (Section \ref{section:motivation}), an important issue for autonomous agents is robustness of the running Bayesian inference processes, even when nodes or edges collapse under situated conditions. ReactiveMP.jl supports in-place model adjustments during the inference procedure as well as handles missing data observations, but it does not export any user-friendly API yet. For the next release of our framework, we aim to export a public API for robust Bayesian inference to simplify the development of Active Inference agents. In addition, our plan is to proceed with further optimisation of the current implementation and improve the scalability of the existing package in real-time on embedded devices. Moreover, the Julia programming language is developing and improving, thus we expect it to be even more efficient in the coming years.

Reactive programming allows us to easily integrate additional features into our new framework. First, we are investigating the possibility to run reactive message passing-based inference in parallel on multiple CPU cores \citep{sarkka_temporal_2020}. RP does not make any assumptions in the underlying data generation process and does not distinguish synchronous and asynchronous data streams. Secondly, reactive systems allow us to integrate local stopping criteria for passing messages. In some scenarios, we may want to stop passing messages based on some pre-specified criterion and simply stop reacting to new observations and save battery life on a portable device or to save computational power for other tasks. Thirdly, our current implementation does not provide extensive debugging tools yet, but it might be crucial to analyse the performance of message passing-based methods step-by-step. We are looking into options to extend the graphical notation and integrate useful debug methods with the new framework to analyze and debug message passing-based inference algorithms visually in real-time and explore their performance characteristics. Since the ReactiveMP.jl package uses reactive observables under the hood, it should be straightforward to ``spy'' on all updates in the model for later or even real-time performance analysis. That should even further simplify the process of model specification as one may change the model in real-time and immediately see the results.

Moreover, we are in a preliminary stage of extending the set of available message passing rules to non-conjugate prior and likelihood pairs \citep{akbayrak_extended_2021}. The generic probabilistic toolboxes in the Julia language like Turing.jl support inference in a broader range of probabilistic models, which is currently not the case for the ReactiveMP.jl. We are working on future message update rules extension and potential integration with other probabilistic frameworks in the Julia community. 

Finally, another interesting future research direction is to decouple the model specification language from the factorisation and the form constraint specification on a variational family of distributions $\mathcal{Q}_B$. This would allow to have a single model $p(s, y)$ and a set of constrained variational families of distributions $\mathcal{Q}_{B_i}$ so we could run and compare different optimisation procedures and their performance simultaneously or trade off computational complexity with accuracy in real-time.

\section{Conclusions}\label{section:conclusions}

In this paper, we presented methods and implementation aspects of a reactive message passing (RMP) framework both for exact and approximate Bayesian inference, based on minimization of a constrained Bethe Free Energy functional. The framework is capable of running hybrid algorithms with BP, EP, EM, and VMP analytical message update rules and supports local factorisation as well as form constraints on the variational family. We implemented an efficient proof-of-concept of RMP in the form of the ReactiveMP.jl package for the Julia programming language, which exports an API for running reactive Bayesian inference, and scales easily to large state space models with hundreds of thousands of unknowns. The inference engine is highly customisable through ad hoc construction of custom nodes, message update rules, approximation methods, and optional modifications to the default computational pipeline. Experimental results for a set of various standard signal processing models showed better performance in comparison to other existing Julia packages and sampling-based methods for Bayesian inference. We believe that the reactive programming approach to message passing-based methods opens a lot of directions for further research and will bring real-time Bayesian inference closer to real-world applications.

\section*{Acknowledgement}

We would like to acknowledge the assistance and their contributions to the source code of the ReactiveMP.jl package from Ismail Senoz, Albert Podusenko, and Bart van Erp. We acknowledge the support and helpful discussions from Thijs van de Laar - one of the creators of the ForneyLab.jl package. We are also grateful to all BIASlab team members for their support during the early stages of the development of this project.

\bibliographystyle{unsrtnat}
\bibliography{main} 

\end{document}